\documentclass[runningheads]{llncs}
\usepackage{graphicx}
\usepackage{arydshln}
\usepackage{caption}
\usepackage{tikz}
\usepackage{comment}
\usepackage{amsmath,amssymb} % define this before the line numbering.
\usepackage{color}

\usepackage[accsupp]{axessibility}  % Improves PDF readability for those with disabilities.

\usepackage{wrapfig}
\usepackage{graphicx}
\usepackage{amsmath}
\usepackage{amssymb}
\usepackage{booktabs}
\usepackage{cases}
\usepackage[ruled,vlined,noend]{algorithm2e}
\usepackage{adjustbox}

\newcommand{\x}{\mathbf{x}}
\newcommand{\w}{\mathbf{w}}
\def\eg{\textit{e.g.}\xspace}
\def\ie{i.e.~}

\usepackage{xr}
\makeatletter
\newcommand*{\addFileDependency}[1]{% argument=file name and extension
  \typeout{(#1)}
  \@addtofilelist{#1}
  \IfFileExists{#1}{}{\typeout{No file #1.}}
}
\makeatother

\newcommand*{\myexternaldocument}[1]{%
    \externaldocument{#1}%
    \addFileDependency{#1.tex}%
    \addFileDependency{#1.aux}%
}
\myexternaldocument{supplementary}

\begin{document}
% \linenumbers
%\pagestyle{headings}
%\mainmatter
%\def\ECCVSubNumber{4899}  % Insert your submission number here

\title{Complementing Brightness Constancy with Deep Networks for Optical Flow Prediction}

%\begin{comment}
%\titlerunning{ECCV-22 submission ID \ECCVSubNumber} 
%\authorrunning{ECCV-22 submission ID \ECCVSubNumber} 
%\author{Anonymous ECCV submission}
%\institute{Paper ID \ECCVSubNumber}
%\end{comment}
%******************

%\begin{comment}
\titlerunning{COMBO: Complementing Brightness Constancy for Optical Flow}
\author{Vincent Le Guen\inst{1,2,3} \and
Clément Rambour\inst{2} \and Nicolas Thome\inst{2,4}}

\authorrunning{Vincent Le Guen, Clément Rambour and Nicolas Thome}

\institute{EDF R\&D, Chatou, France \\
 \and
Conservatoire National des Arts et Métiers, CEDRIC, Paris, France \\ \and
SINCLAIR AI Lab, Palaiseau, France
 \and
Sorbonne Université, CNRS, ISIR, F-75005 Paris, France
}

%\end{comment}
%******************
\maketitle

\begin{abstract}
State-of-the-art methods for optical flow estimation rely on deep learning, which require complex sequential training schemes to reach optimal performances on real-world data. In this work, we introduce the COMBO deep network that explicitly exploits the brightness constancy (BC) model used in traditional methods. Since BC is an approximate physical model violated in several situations, we propose to train a physically-constrained network complemented with a data-driven network. We introduce a  unique and meaningful flow decomposition between the physical prior and the data-driven complement, including an uncertainty quantification of the BC model. We derive a joint training scheme for learning the different components of the decomposition ensuring an optimal cooperation, in a supervised but also in a semi-supervised context. Experiments show that COMBO can improve performances over state-of-the-art supervised networks, \eg RAFT, reaching state-of-the-art results on several benchmarks. We highlight how COMBO can leverage the BC model and adapt to its limitations. Finally, we show that our semi-supervised method can significantly simplify the training procedure.
\end{abstract}

\begin{figure*}[!h]
    \centering
    \includegraphics[width=\linewidth]{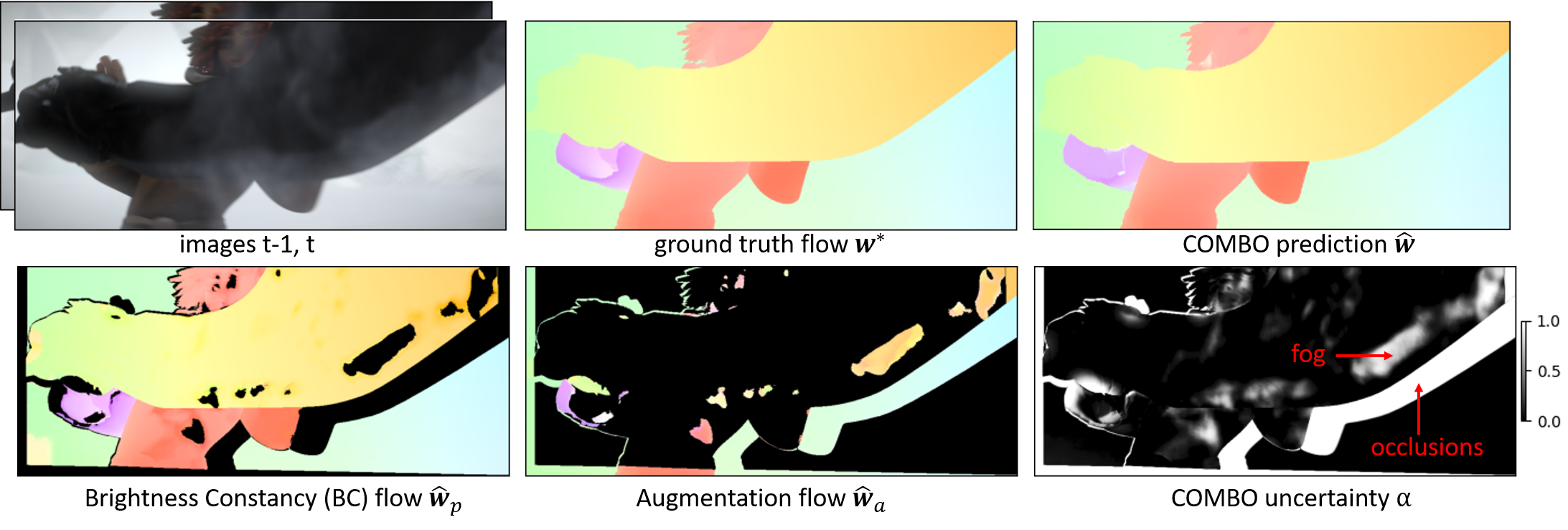}
   \caption{\textbf{Proposed COMBO model for flow estimation} between two images $I_1$ and $I_2$. We explicitly leverage the brightness constancy (BC) assumption in Eq.~(\ref{eq:flot}) for estimating the physical flow prediction $\hat{\w}_p$. Since BC is an approximate model violated in several usual situations, COMBO learns an uncertainty map $\hat{\alpha}$ specifying areas where $\hat{\w}_p$ is relevant, and an augmentation data-driven flow $\hat{\w}_a$ to compensate for the BC violations (in this example, related to fog and occlusions). The final COMBO prediction combining $(\hat{\w}_a, \hat{\w}_p, \hat{\alpha})$ is both physically-regularized and accurate.}
    \label{fig:intro}
\end{figure*}%

\section{Introduction}

Optical flow estimation is a classical problem in computer vision, consisting in computing the per-pixel motion between video frames. This is a core visual task useful in many applications, such as video compression \cite{agustsson2020scale}, medical imaging \cite{balakrishnan2018unsupervised} or object tracking \cite{corpetti2002dense}. Yet, this remains a particularly challenging task for real-world scenes, especially in presence of fast-moving objects, occlusions, changes of illumination or textureless surfaces.

Traditional model-based estimation methods \cite{lucas1981iterative,horn1981determining} assume that the intensity of pixels is conserved during motion $\w$:  $I_{t-1}(\x) = I_t(\x+\w)$, which is known as the \textit{brightness constancy} (BC) model. Linearizing this constraint for small motion leads to the famous optical flow partial differential equation (PDE) solved by variational methods since the seminal work of Horn and Schunk \cite{horn1981determining}: 
\begin{equation}
    \frac{\partial I}{\partial t} (t,\mathbf{x}) + \w(t,\mathbf{x}) \cdot \nabla I (t,\mathbf{x}) = 0
    \label{eq:flot}
\end{equation}

However this physical model for optical flow is only a rather coarse approximation of the reality. The BC assumption is violated in many situations: in presence of occlusions, global or local illumination changes, specular reflexions, or complex natural situations such as fog. The correspondence problem is also ambiguous for textureless surfaces. To solve this ill-posed inverse problem, this is necessary to inject prior knowledge about the flow, such as spatial smoothness, sparsity or small total variation.

In contrast to these traditional model-based approaches, deep neural networks have become state-of-the-art for learning optical flow in a pure data-driven fashion \cite{dosovitskiy2015flownet,ilg2017flownet,ranjan2017optical,sun2018pwc,hui2018liteflownet,yang2019volumetric,raft}. However, since flow labelling on real images is expensive, supervised deep learning methods mostly use complex curriculum training schemes on synthetic datasets to progressively adapt to the complexity of real-world scenes.

%This assumption is violated in the following cases:
%\begin{itemize}
%\item \textbf{Occlusions:} when a pixel becomes occluded in frame 2, we loose information to reconstruct it in frame 1. It causes ghosting effects.
%\item \textbf{Global or local illumination changes:}
%\item \textbf{Specular reflexions:}
%\item \textbf{Textureless regions:} causes ambiguities for matching pixels
%\end{itemize}

%More recently, deep learning approaches have proposed learning optical flow end-to-end and have become state-of-the-art. Two classes of methods exits: supervised and unsupervised. In the supervised context, DL methods do not exploit the brightness constancy hypothesis anymore, or indirectly (through the computation of a cost volume). They rely on large datasets of annotated image pairs (often on synthetic datasets). 

More recently, unsupervised deep learning approaches are closer in spirit to traditional approaches \cite{jason2016back,ren2017unsupervised,meister2018unflow,liu2020learning,jonschkowski2020matters,stone2021smurf}. Without ground-truth labels, they rely on a photometric reconstruction loss based on the BC assumption, and additional regularization losses like spatial smoothness. They also use of specific modules, \eg occlusion detection, to identify images areas where the BC in Eq.~(\ref{eq:flot}) is violated.

%\textcolor{red}{Remove this paragraph?} Combining model-based (MB) methods with modern machine learning (ML), in particular leveraging physical ordinary or partial differential equations (ODE/PDE), is a thriving subject nowadays. Although most methods assume fully-specified physical models, a few recent hybrid MB/ML models have explored how to exploit \textit{incomplete} physical models for forecasting dynamical systems. Incomplete models are widespread in natural sciences, \eg for modelling complex phenomena like climate.

In this work, we introduce a hybrid model-based machine learning (MB/ML) model dedicated to optical flow estimation, which COMplements Brightness constancy with deep networks for accurate Optical flow prediction (COMBO). As illustrated in Figure \ref{fig:intro}, COMBO  decomposes the estimation process into a physical flow based on the simplified Brightness Constancy (BC) hypothesis and a data-driven augmentation for compensating for the limitations of the physical model. COMBO also simultaneously learns an uncertainty model of the brightness constancy, useful for fusing the two flow branches into an accurate flow estimation. Importantly, the two branches are learned jointly and a principled optimization ensures an optimal cooperation between them.

Our contributions are the following:
\begin{itemize}
    \item We propose a principled way to regularize supervised flow models with an explicit exploitation of the BC assumption. We introduce with COMBO a meaningful and unique flow decomposition%Our hybrid MB/ML model, termed AugmFlow, decomposes the flow 
    ~into a physical part optimizing the BC and an augmentation part compensating for its limitations.

    \item We propose to jointly learn the physical, augmented flows and a uncertainty measure of the BC validity, generalizing occlusion detection approaches or robust photometric losses. It also enables to exploit the model in semi-supervised contexts for non-annotated pixels where we know that the BC assumption is satisfied.
    
    \item To learn the COMBO model, we define and compute a fine-grained supervision by decomposing the ground-truth flow into a BC flow, a residual flow and an uncertainty map.  
 
    \item Experiments (section \ref{sec:expes}) show that when included into a state-of-the-art supervised network, \ie RAFT, COMBO can improve performances consistently over datasets and training curriculum steps. COMBO also reaches state-of-the-art performances on several benchmarks, and we show that the semi-supervised scheme can be leveraged to grandly simplify the training curriculum. We also provide detailed ablations and qualitative analyses attesting the benefits of our principled decomposition framework.
\end{itemize}

\section{Related work}

\paragraph{\textbf{Traditional optical flow}}
Since the seminal works of Lucas-Kanade \cite{lucas1981iterative} and Horn-Schunck \cite{horn1981determining}, optical flow estimation is traditionally casted as an optimization problem over the space of  motion fields between a pair of images \cite{memin1998dense,wedel2009structure,brox2010large}. As mentionned in introduction, the cost function is usually based on the conservation assumption of an image descriptor during motion, \ie %the most common being 
the \textit{brightness constancy} (BC).%\textcolor{red}{REPTITION INTRO}
~Since the number of optical flow variables is twice the number of observations, the search problem is underconstrained. Existing methods typically add spatial smoothness constraints on the flow to regularize the problem, enabling to propagate information from non-ambiguous to ambiguous pixels, \eg occluded. 
%Local approaches \cite{lucas1981iterative} estimate a per-pixel flow by solving the brightness constancy equations on small overlapping patches. Global approaches \cite{horn1981determining,sun2010secrets} solve for the whole flow map by optimizing an energy combining the similarity of matched pixels and a spatial smoothness prior; global regularization enables to propagate information from non-ambiguous to ambiguous pixels, \eg occluded. 
Since, progress has been made for handling the limitations of the brightness constancy with more robust matching loss, such as the gradient constancy \cite{brox2004high}, the structural similarity (SSIM), the Charbonnier loss \cite{bruhn2005lucas}, the census loss \cite{meister2018unflow} or involving descriptors learned by deep neural networks \cite{bailer2017cnn,xu2017accurate}. However, these traditional methods are limited for handling large motion and often suffer from a high computational cost.

\paragraph{\textbf{Deep supervised methods}} Deep neural networks have proven effective for directly learning the optical flow end-to-end from a labelled dataset \cite{dosovitskiy2015flownet,ilg2017flownet,ranjan2017optical,sun2018pwc,hui2018liteflownet,raft,jiang2021learning}. Although some methods borrow ideas from traditional methods such as cost volume processing \cite{yang2019volumetric} and pyramidal coarse-to-fine warping \cite{ranjan2017optical}, they do not rely anymore on the conservation of an image descriptor. One of the current state-of-the-art architectures is the  Recurrent All-Pairs Field Transforms (RAFT) \cite{raft}. This model exploits the correlation volume between all pairs of pixels and iteratively refines the flow with a recurrent neural network. Since obtaining ground-truth flow annotations is very expensive, supervised methods mostly train on synthetic datasets. This leads to a non-negligible generalization gap when transferring to real-world datasets. Consequently, complex curriculum learning schemes are needed to progressively adapt the learned optical flow from synthetic data to real scenarios \cite{raft,sun2021autoflow}.

%Neural networks have been trained to directly predict optical flow between a pair of frames, side-stepping the optimization problem completely. Coarse-to-fine processing has emerged as a popular ingredient in many recent works [42,50,22,23,24,49,20,8,52]. In contrast, our method maintains and

%The current best network architecture is thl. It follows classical work that breaks with the coarse-to-fine assump- tion [4, 26, 36] and computes the cost volume between all pairs of pixels and uses that information to iteratively refine the flow field [29]. 

%RAFT: use cost volume, indirect exploitation of brightness constancy

\paragraph{\textbf{Deep unsupervised methods}} 
More recently, unsupervised approaches proved to outperform traditional methods even without flow labels \cite{jason2016back,ren2017unsupervised,meister2018unflow,liu2020learning,jonschkowski2020matters,stone2021smurf}. They revisit the brightness constancy assumption by warping the target image with the estimated flow and optimizing a photometric loss. However they still lag behind supervised methods and require additional mechanisms for overcoming the limitations of the BC assumption, \eg occlusion reasoning \cite{meister2018unflow,wang2018occlusion} or self-supervision \cite{liu2019ddflow,liu2019selflow,jonschkowski2020matters}. 
%More recently, unsupervised methods overcome the need of labelled data by revisiting the traditional ideas of photometric constancy with deep learning . 
%Later works highlighted the crucial necessity to exclude occluded pixels from the photometric loss 
Occlusions were traditionally treated as outliers in a robust estimation setting \cite{brox2004high}, or estimated with a forward-backward consistency check \cite{alvarez2007symmetrical,meister2018unflow}. More recent approaches jointly learn occlusion maps with convolutional neural networks in a supervised way (\ie with ground-truth occlusions) \cite{mayer2016large,ilg2017flownet,hur2019iterative} or unsupervised way \cite{zhao2020maskflownet,godet2021starflow,janai2018unsupervised}. Nonetheless, most of these methods focus on specific cases of deviations from the brightness constancy assumption, \eg occlusions or illumination changes. Our approach is more general and learns to compensate for any failure case of this simplified assumption. Besides, we learn from supervised flow data a confidence model of the BC assumption, which enables to learn our model in a semi-supervised setting \cite{lai2017semi,yan2020optical}.

\paragraph{\textbf{Hybrid MB/ML}}

Combining model-based (MB) and machine learning (ML) models is a long-standing subject  \cite{psichogios1992hybrid,thompson1994modeling,rico1994continuous} that yet remains widely open nowadays. Leveraging physical knowledge enables to design machine learning models that learn from less data, and offer better generalization while conserving physical plausibility. Many ideas were explored, such as imposing soft physical constraints \cite{raissi2018deep,sirignano2018dgm,lu2019deeponet,li2020fourier,wang2021learning} in the training loss function or hard constraints in the network architectures \cite{de2017deep,daw2020physics,leguen-phydnet,greydanus2019hamiltonian,lutter2019deep,mohan2020embedding}. However, most existing approaches assume a fully-known prior physical model. Very few works have investigated how to exploit incomplete physical models with deep neural networks \cite{long2018hybridnet,neural20,saha2020phicnet,linial2020generative,leguen2021augmenting}. Besides, exploiting incomplete physical knowledge in a deep hybrid model has never been explored for optical flow to the best of our knowledge. 

\section{COMBO model for optical flow}

\begin{figure*}
    \centering
    \includegraphics[width=\linewidth]{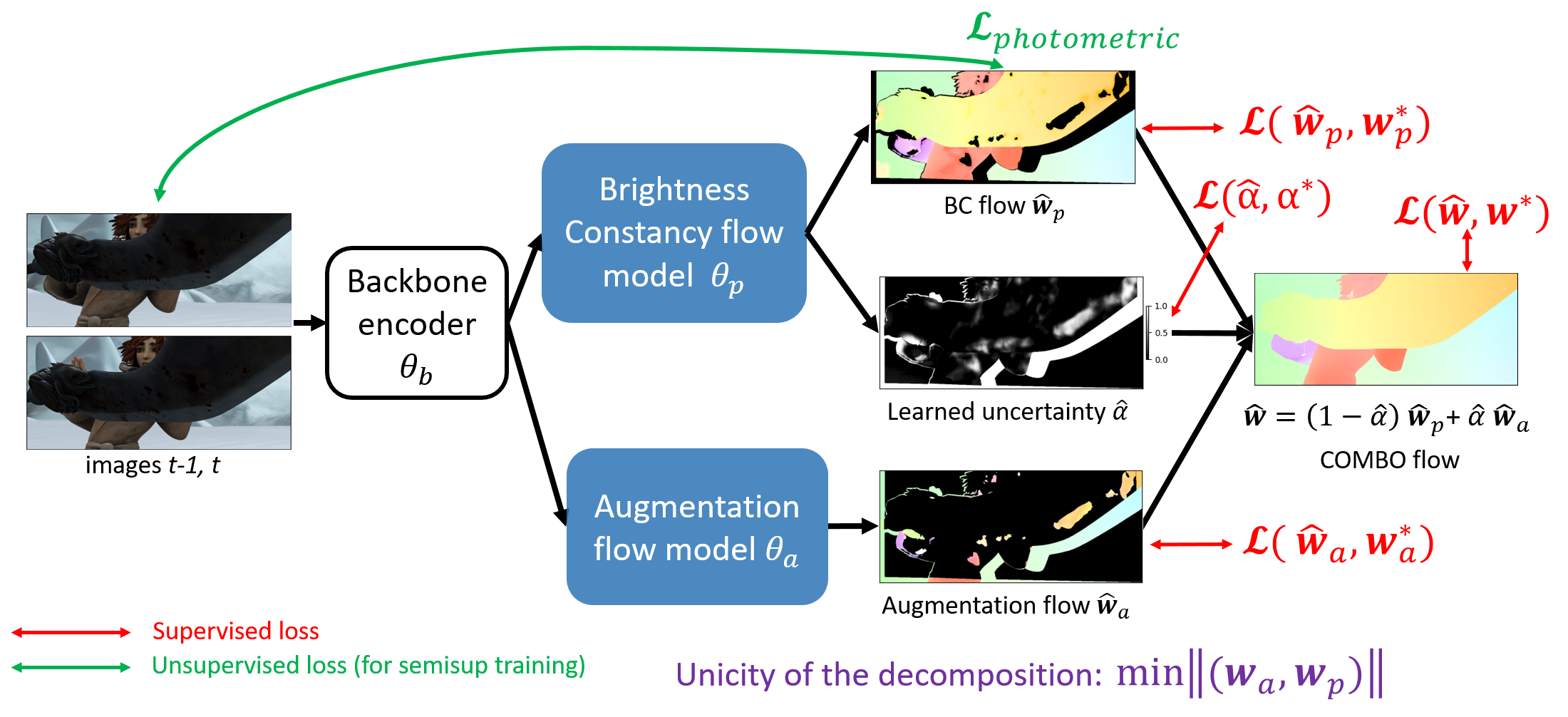}
    \caption{Our proposed COMBO model for optical flow estimation. Given a pair of input frames ($I_{t-1}, I_t$), COMBO first extract convolutional features. The model is composed of three branches that compute an estimate for the physical flow $\hat{\w}_p$, the augmentation flow $\hat{\w}_a$ and the uncertainty map  $\hat{\alpha}$.}
    \label{fig:model_overview}
\end{figure*}

Given a pair of frames $I_{t-1}$ and $I_t \in  \mathbb{R}^{W \times H \times 3}$, the optical flow task consists in estimating the dense motion field $\w = (u,v) \in \mathbb{R}^{W \times H \times 2}$ that maps each pixel from the source image $I_{t-1}$ to its corresponding relative location in the target image $I_t$.

We introduce COMBO, a deep learning model that explicitly exploits the brightness constancy (BC) assumption and learns the complementary information to compensate for the BC modeling errors.

The COMBO model is depicted in Figure \ref{fig:model_overview}. The model takes a pair of input frames $(I_{t-1},I_t)$ and first encodes them into a feature space. From this common representation space, the rationale of COMBO is to decompose the dense flow $\hat{\w}$ using 3 main quantities: a physical term $\hat{\w}_p \in \mathbb{R}^{W \times H \times 2}$ fulfilling the BC assumption, an augmentation term $\hat{\w}_a \in \mathbb{R}^{W \times H \times 2}$ dedicated to capture the remaining information for accurate flow prediction. For each pixel, our decomposition also involves an uncertainty map $\hat{\alpha} \in [0,1]^{W \times H}$ representing the violation of the BC assumption at each pixel. 

In section \ref{sec:augmentation}, we present our proposed BC augmentation framework, ensuring a unique and meaningful decomposition of $\hat{\w}$ between the physical $\hat{\w}_p$, augmented $\hat{\w}_a$ and uncertainty $\hat{\alpha}$ terms. From this formulation, we derive in section \ref{sec:training} a training scheme to learn the model parameters, including a supervised but also a semi-supervised variant leveraging the learned BC confidence $\hat{\alpha}$.

\subsection{BC-augmented flow decomposition}
\label{sec:augmentation}

We consider the flow $\w(\mathbf{x})$ representing the displacement of pixel $\mathbf{x}=(x,y)$ from $I_{t-1}(\mathbf{x})$ to $I_t(\mathbf{x})$. Let us define the brightness constancy divergence as  $\mathcal{L}_{BC}(\x,\w): = \ell(I_{t-1}(\x)-I_t(\x+\w))$   measured with the photometric loss $\ell$ (\eg the L1 loss). In COMBO, we propose to decompose the ground truth flow $\w^*$ between a physical flow $\w_p^*$, an augmentation flow $\w_a^*$ and an uncertainty map $\alpha^*$: 
\begin{equation}
   \w^*(\x)= (1-\alpha^*(\x)) ~ \w_p^*(\x) + \alpha^*(\x) ~ \w_a^*(\x).
   \label{eq:decompo}
\end{equation}

Since the decomposition in Eq.~(\ref{eq:decompo}) is not necessarily unique, we define the COMBO decomposition ($\w_p^*, \w_a^*, \alpha^*$) as the solution of the following constrained optimization problem:

\begin{align}
\underset{\w_p, \w_a}{\min} ~~~  \left\Vert  (\w_a,\w_p)  \right\Vert ~~~~~~~ \text{subject to: ~~~~~~}     \label{eq:optpb} \\
\begin{cases}
(1-\alpha^*(\x)) ~ \w_p(\x) + \alpha(\x) ~ \w_a(\x) = \w^*(\x)  \\
 (1-\alpha^*(\x))~ | I_1(\x) - I_2(\x+\w_p(\x))  | = 0  \nonumber   \\
 \alpha^*(\x) = \sigma \left( |I_1(\x)-I_2(\x+\w^*(\x))| \right).   \\
\end{cases}
\end{align}

\paragraph*{\textbf{Theoretical properties of the decomposition:}} Under these constraints, the decomposition problem in Eq.~(\ref{eq:optpb}) admits a unique solution  $(\w_p^*,\w_a^*,\alpha^*)$ (we detail the proof in supplementary \ref{sup:proof}). The uniqueness in decomposition of a given flow into the three $(\w_p^*,\w_a^*, \alpha^*)$ components is important since it ensures a well posed problem for learning the different terms.

Beyond its sound mathematical formulation, the decomposition in Eq.~(\ref{eq:optpb}) is also meaningful. It ensures to properly and explicitly exploit the BC model and to overcome its limitations to represent real data in diverse and complex situations.
We detail now each component of this decomposition problem:

\paragraph*{\textbf{Brightness constancy flow $\hat{\w}_{p}(\mathbf{x})$}:} We define the brightness constancy flow $\hat{\w}_p$ as a minimizer of the BC condition, similarly to traditional and unsupervised methods. The BC is a very useful constraint not fully exploited in deep supervised methods. However, as mentioned in introduction, this BC assumption is often violated in real data, \eg in case of occlusions, illumination changes, textureless surfaces, \textit{etc}. When directly minimizing $\mathcal{L}_{BC}(\x,\w) = \ell(I_{t-1}(\x)-I_t(\x+\w))$, for a given pixel $\x$ in the source image, there may exist multiple possible matches in the target image satisfying the BC constraint. Therefore, the BC constraint is only applied to $\w_p$ in Eq.~(\ref{eq:optpb}), not to the final flow vector $\w^*$, and is weighted by the BC confidence measure ($1-\alpha^*(\x)$). This is done on purpose, since the BC constraint is not always verified when training a model from ground truth supervision, making this term possibly conflicting with the supervised objective. We verify experimentally that a naive incorporation of the BC loss during training is not effective. In contrast, COMBO exploits an augmented flow to jointly incorporate the BC knowledge with an adaptive compensation for its violation.

\paragraph*{\textbf{BC augmented flow $\hat{\w}_{a}(\mathbf{x})$}:} 
In contrast to traditional and unsupervised methods, we do not leverage prior knowledge on the flow but instead rely on a data-driven augmentation $\hat{\w}_a$ to learn how to optimally compensate for the simplified model, as investigated by several recent augmented physical models \cite{long2018hybridnet,wang2019integrating,neural20,saha2020phicnet,linial2020generative,leguen2021augmenting}. To ensure a unique decomposition in Eq.~(\ref{eq:optpb}), we minimize the norm of the concatenated vector $(\hat{\w}_a,\hat{\w}_p)$. The rationale is to compensate with  the minimal correction $\Vert \hat{\w}_a \Vert$ from the brightness constancy assumption. This least-action augmentation also prevents $\hat{\w}_p(\x)$ from being too large: the BC flow is enforced to be as close as possible to the ground-truth flow $\w^*$. Minimizing the norm of $\hat{\w}_p$ also prevents degenerate cases with possibly several admissible $\hat{\w}_p$ equidistant to $\w^*$ (see supplementary \ref{sup:proof} for a discussion).

\paragraph*{\textbf{BC uncertainty $\alpha^*(\mathbf{x})$}:} 
We weight the decomposition between $\hat{\w}_p$ and $\hat{\w}_a$ in Eq.~(\ref{eq:optpb}) with the uncertainty map  $\alpha^*(\mathbf{x}) = \sigma( \mathcal{L}_{BC}(\x,\w^*) )$ quantifying the per-pixel validity of the BC assumption, where $\sigma$ is a nonlinear function\footnote{We choose in this work the sigmoid function centered at 0.5 for image pixels in the range $[0;1]$.} ensuring that the uncertainty values spread over $[0;1]$. When $\alpha^*(\x)=0$, the BC assumption is verified; in that case, our decomposition reduces to the physical flow $\hat{\w}(\x) = \hat{\w}_p(\x)$ as in traditional methods. On the contrary, when the BC is violated ($\alpha^*(\x)> 0$), the errors of the brightness constancy modelling are compensated by the data-driven model $\hat{\w}_a$. This leads to a meaningful decomposition between $\hat{\w}_p$ and $\hat{\w}_a$.

\subsection{Training}
\label{sec:training}

In this section, we propose to jointly learn the three quantities $\w_p^*$, $\w_a^*$ and $\alpha^*$ of the flow decomposition Eq.~(\ref{eq:decompo}) with a deep neural architecture named COMBO. We introduce a practical learning framework to solve the decomposition problem in Eq.~(\ref{eq:optpb}) in the supervised and semi-supervised settings.  Given a dataset of labelled image pairs $\mathcal{D}_{sup} = \left\{  (I_{t-1},I_t,\w^*)_i  \right\}_{i=1}^{N_{sup}}$ and unlabelled image pairs $\mathcal{D}_{unsup} = \left\{  (I_{t-1},I_t)_i  \right\}_{i=1}^{N_{unsup}}$, the goal is to learn the parameters $\theta = \left\{\theta_b, \theta_p, \theta_a \right\}$ of the COMBO model depicted in Figure \ref{fig:model_overview}, where $\theta_b$ are the backbone encoder parameters, $\theta_p$ the parameters of the BC flow model, $\theta_a$ are the parameters of the augmented flow model. 
 
The general loss for learning the COMBO model is the following:
\begin{multline}
     \mathcal{L}(\mathcal{D}, \theta) =  \lambda_{p}~ \Vert \hat{\w}_p-\w_p^* \Vert_2^2 +  \lambda_{a}~ \Vert \hat{\w}_a-\w_a^* \Vert_2^2 + \lambda_{total}~ \Vert \hat{\w}-\w^* \Vert_2^2 \\
      + \lambda_{photo} ~   \mathcal{L}_{photo}(\mathcal{D},\theta)  
       + \lambda_{w} ~ \Vert (\hat{\w}_a,\hat{\w}_p) \Vert^2   + \lambda_{\alpha} ~  \mathcal{L}_{\alpha}(\mathcal{D},\theta),
       \label{eq:loss_sup}
\end{multline}
where $\hat{\w}(\x) = (1-\hat{\alpha}(\x))~ \hat{\w}_p(\x) + \hat{\alpha}(\x) ~ \hat{\w}_a(\x)$ is the flow predicted by COMBO.

The total loss in Eq.~(\ref{eq:loss_sup}) is composed of supervised losses on the final flow $\hat{\w}$, the BC flow $\hat{\w}_p$ and the augmentation flow   $\hat{\w}_a$. In addition to the supervised loss on $\hat{\w}_p$, we also use a photometric loss similarly to traditional and unsupervised methods. We use the L1 loss weighted by the certainty measure $(1-\alpha^*(\x))$: 
\begin{equation}
    \mathcal{L}_{photo}(\mathcal{D},\theta) = \sum_{I_1, I_2 \in \mathcal{D}} \sum_{\x} \; (1-\alpha^*(\x))  ~ |I_1(\x)-I_2(\x+\hat{\w}_p)|_1.
    \label{eq:l_photo}
\end{equation}

With the L1 loss, the physical branch of COMBO directly models the brightness conservation at the pixel level and the learned uncertainty captures the limitations of this hypothesis. Other photometric losses more robust to the brightness constancy limitations could have been chosen, \eg the Charbonnier \cite{bruhn2005lucas}, SSIM \cite{jonschkowski2020matters} or census loss \cite{meister2018unflow}, that would produce a different physical flow. Our augmentation framework can seamlessly adapt to the level of approximation of the BC model at each pixel. Note that the photometric loss is complementary with the supervised loss $\Vert \hat{\w}_p-\w_p^* \Vert_2^2$: a small error on the predicted flow  $\hat{\w}_p$ results in a small endpoint error (EPE) but can have a large photometric error  if  $\hat{\w}_p$ leads to a large photometric change in the target image.

Finally, we define the uncertainty loss as:
\begin{equation}
   \mathcal{L}_{\alpha}(\mathcal{D}_{sup},\theta) = \sum_{I_1, I_2 \in \mathcal{D}_{sup}}  \sum_{\x} \; \Vert \hat{\alpha}(\x) -  \sigma( \mathcal{L}_{BC}(\x,\w^*) )  \Vert_2^2. 
\end{equation}

\noindent \textbf{Curriculum learning for the BC uncertainty:}
In the COMBO decomposition  $\hat{\w}(\x) = (1-\hat{\alpha}(\x))~ \hat{\w}_p(\x) + \hat{\alpha}(\x) ~ \hat{\w}_a(\x)$, a bad estimate of the uncertainty $\alpha$ could be harmful for learning $\hat{\w}_p$ and $\hat{\w}_a$. Therefore, similarly to the \textit{teacher forcing} strategy in sequential models, we choose a scheduled sampling strategy \cite{bengio2015scheduled} where we use the ground truth $\alpha^*$ with high probability at the beginning of learning, and decrease progressively this probability towards 0 to rely more and more on the prediction $\hat{\alpha}$ (see Algorithm \ref{alg:optim}).

\subsubsection{Supervised learning:}

In the supervised learning setting, we only exploit the supervised set $\mathcal{D}_{sup}$. We minimize the total loss in Eq.~(\ref{eq:loss_sup}) by using the fine-grained supervision on $(\w_p^*,\w_a^*, \alpha^*)$  that we have created (see supplementary \ref{sup:supervision} for examples of tuples  $(\w_p^*,\w_a^*, \alpha^*)$ for each dataset).

\subsubsection{Semi-supervised learning:}

COMBO can be also favorably used in a semi-supervised training setting, which consists in exploiting the information of non-annotated images in $\mathcal{D}_{unsup}$ in addition to the annotated frames in $\mathcal{D}_{sup}$. This is an important context in practice since labelling flow images is expensive at scale (in general $N_{unsup} \gg N_{sup}$). Leveraging unlabelled data directly on the target dataset is an appealing way for unsupervised domain adaptation.

When learning in semi-supervised mode (see Algorithm \ref{alg:optim}), we mix in a mini-batch labelled image pairs from $\mathcal{D}_{sup}$ and unlabelled image pairs from $\mathcal{D}_{unsup}$. For labelled images, we again minimize the supervised loss $\mathcal{L}$ defined in Eq.~(\ref{eq:loss_sup}). For unlabelled frames, we refine the estimation of $\hat{\w}_p$ by exploiting the photometric loss on pixels that are likely to satisfy the BC constraint (measured by the uncertainty map $\alpha$ learned with ground-truth data). We only minimize in that case the photometric loss $\mathcal{L}_{photo}(\mathcal{D}_{unsup},\theta)$, which corresponds to the particular case of Eq.~(\ref{eq:loss_sup}) by setting $\lambda_{photo}=1$ and $\lambda_{p}=\lambda_a=\lambda_{total}=\lambda_w=\lambda_{\alpha}=0$. Importantly, for unlabelled images, we block the gradient flow in the uncertainty branch $\alpha$ since $\alpha$ can only be estimated in supervised mode. 

\begin{algorithm}
\SetAlgoLined
Parameters: $\lambda_{p}, \lambda_a, \lambda_{total}, \lambda_{photo}, \lambda_w, \lambda_{\alpha} \geq 0,  \tau >0$ \;
\For{epoch = $1:N_{epochs}$} {
\For{each batch $b=(b_{sup},b_{unsup})$}{

\textbf{If} $Rand(0,1) <  1 - epoch/50$ \textbf{Then}:  $~~~~\#$ \texttt{scheduled sampling}

$\hat{\w}(\x) = (1-\alpha^*(\x))~ \hat{\w}_p(\x) + \alpha^*(\x) ~ \hat{\w}_a(\x)$\\
\textbf{Else:}\\  $\hat{\w}(\x) = (1-\hat{\alpha}(\x))~ \hat{\w}_p(\x) + \hat{\alpha}(\x) ~ \hat{\w}_a(\x)$
\\
Compute $\mathcal{L}_{sup} = \mathcal{L}(b_{sup},\theta, \lambda_{sup}, \lambda_{photo},\lambda_w, \lambda_{\alpha})$ with Eq.~(\ref{eq:loss_sup})  \\
Compute $\mathcal{L}_{unsup} = \mathcal{L}_{photo}(b_{unsup}, \theta)$ \\
$\theta_{j+1} = \theta_j - \tau ~ \nabla (\mathcal{L}_{sup}+ \mathcal{L}_{unsup})  \;$
}
}
  \caption{\label{alg:optim} COMBO optimization:}
\end{algorithm}

\begin{table}[b!]
   \centering
   \begin{adjustbox}{max width=\linewidth}
    \begin{tabular}{cc|ccccc}
    \toprule
Stage &  Method &  Chairs & \multicolumn{2}{c}{Sintel-test-resplit}  & \multicolumn{2}{c}{KITTI-test-resplit}   \\
\cmidrule(r){4-5}  \cmidrule(r){6-7}
 ~ & ~ & ~ & Clean & Final & Fl-epe & Fl-all
\\ \midrule
   C &  RAFT   & 0.82 & 1.11  & 1.76  & 10.7  & 39.7    \\ 
    C & COMBO  & \textbf{0.74} & \textbf{1.04} & \textbf{1.58}  & \textbf{10.2} & \textbf{39.3} \\
%    C & COMBO2 & 0.71 & 1.07 & 1.61 & 10.9 & 42.1 \\
    \midrule
  C+T &  RAFT  & 1.15 & 0.82  & 1.17  & 5.67 & 17.9 \\ 
  C+T & COMBO  & \textbf{1.09} & \textbf{0.77} & \textbf{0.96} & \textbf{5.46} & \textbf{17.3} \\
    \midrule
 C+T+ &   RAFT   & 1.23 & 0.57  &  0.76 & 1.79  & 7.12  \\
 S+H+K & COMBO  & \textbf{1.16} & \textbf{0.56}  & \textbf{0.71} & \textbf{1.75} & \textbf{6.58}   \\
    \bottomrule
    \end{tabular}
    \end{adjustbox}
    \caption{Performances of COMBO compared to the RAFT model (run with online code). COMBO consistently outperforms RAFT on the three training stages, illustrating the relevance of leveraging the BC physical model and complementing its failure cases.}
    \label{tab:superv1}
\end{table}

\section{Experiments}
\label{sec:expes}

We evaluate and compare COMBO to recent state-of-the-art models on the optical flow datasets FlyingChairs \cite{dosovitskiy2015flownet}, MPI Sintel \cite{butler2012naturalistic} and KITTI-2015 \cite{geiger2013vision}, in the supervised and semi-supervised contexts. For model analysis, we perform a train/test resplit of the Sintel and KITTI training datasets, as done by \cite{liu2020learning}. Our code is available at \texttt{https://github.com/vincent-leguen/COMBO}.

%\subsection{Neural network architecture}
\subsection{Experimental setup}
\label{sec:nn_impl}

 We adopt for the backbone encoder and flow network the RAFT architecture \cite{raft}, which is one of the current state-of-the-art methods for supervised optical flow. The encoder is composed of convolutional layers that extract features from images $I_{t-1}$ and $I_t$ at resolution $1/8$. Then visual similarity is modelled by constructing a 4D correlation volume that computes matching costs for all possible displacements. The optical flow branch is a gated recurrent unit (GRU) that progressively refines the flow from an initial estimate, with lookups on the correlation volume. The final flow is the sum of residual refinements from the GRU, upsampled at full resolution.

In the COMBO model, the encoder and correlation volume are shared, and each branch has its own GRU initialized from zero. The uncertainty branch is simply adapted to provide a unique output channel. We give details on the network architectures and hyperparameters in supplementary \ref{sup:implem}.  The code of COMBO will be released if accepted. Note that our method is agnostic to the optical flow backbone. Any other deep architecture than RAFT could be used for computing the physical, augmentation flow and the uncertainty map.

\subsection{Supervised state-of-the-art comparison}

Following the supervised learning curriculum of prior works \cite{zhao2020maskflownet,sun2019models,raft,jiang2021learning}, we pretrain our models successively on the synthetic datasets FlyingChairs \cite{dosovitskiy2015flownet} and FlyingThings3D \cite{mayer2016large}. We then finetune on Sintel \cite{butler2012naturalistic} by combining data from Sintel, KITTI-2015 and HD1K \cite{kondermann2016hci}. We finally finetune on KITTI-2015 with only data from KITTI.

\begin{table}[t!]
   \centering
   \begin{adjustbox}{max width=\linewidth}
    \begin{tabular}{cc|cccccc}
    \toprule
 Stage  &  Method &  Chairs & \multicolumn{2}{c}{Sintel (train)}  & \multicolumn{2}{c}{KITTI-15 (train)}  \\
\cmidrule(r){4-5}  \cmidrule(r){6-7} 
  & ~ & ~ & Clean & Final & Fl-epe & Fl-all 
\\ \midrule
   C &  RAFT \cite{raft}  & 0.82 & 2.26 & 4.52 & 10.67 & 39.72   \\ 
   C & GMA \cite{jiang2021learning}  & 0.79 & 2.32 & \textbf{4.10} & 10.32 & \textbf{36.91} \\
    C & COMBO  & \textbf{0.74} & \textbf{2.19} & 4.37  & \textbf{10.19} & 39.02  \\
%    C & COMBO2 & 0.71 & 2.30 & 4.59 & 10.14 & 39.79 \\
    \midrule
    C+T & LiteFlowNet \cite{ilg2017flownet} &  - & 2.48 & 4.04 & 10.39 & 28.5 \\
      C+T & PWC-Net \cite{sun2018pwc} & - & 2.55 & 3.93 & 10.35 & 33.7  \\
  C+T & VCN \cite{yang2019volumetric} & - & 2.21 & 3.68 & 8.36 & 25.1 \\    
  C+T & MaskFlowNet \cite{zhao2020maskflownet} & - & 2.25 & 3.61 & - & 23.1 \\    
  C+T &  RAFT \cite{raft}  & 1.15 & 1.43 & 2.71 & 5.01 & 17.5  \\ 
 C+T  &  GMA \cite{jiang2021learning}  & 1.19 & \textbf{1.31} & 2.73 & \textbf{4.69} & 17.1 \\
  C+T & COMBO  & \textbf{1.09} & \textbf{1.31} & \textbf{2.58} & \textbf{4.69} & \textbf{16.5}   \\
    \bottomrule
    \end{tabular}
    \end{adjustbox}
    \caption{State-of-the-art performance comparison on FlyingChairs, Sintel-train and KITTI-train, when learning on FlyingChairs (C) and FlyingChairs+FlyingThings (C+T). Scores are reported from corresponding papers.}
    \label{tab:superv2}
\end{table}

In Table \ref{tab:superv1}, we compare the endpoint error (EPE) of COMBO with RAFT \cite{raft}, by running the code from the authors. Since we rely on a RAFT architecture, the results in Table \ref{tab:superv1} directly evaluate the impact of the proposed augmented model and learning scheme. We see that COMBO consistently outperforms RAFT on FlyingChairs, Sintel-test-resplit and KITTI-test-resplit from all stages of the curriculum. This confirms the relevance of leveraging the BC constraint for regularizing supervised models, and the use of the augmentation for compensating its violations. 

In Table \ref{tab:superv2}, we compare COMBO to competitive supervised optical flow baselines, from the stages FlyingChairs (C) and FlyingChairs+FlyingThings (C+T). When generalizing to Sintel-train and KITTI-train, COMBO outperforms RAFT \cite{raft} in all cases, and is superior or equivalent to the recent GMA
model \cite{jiang2021learning} in 6/8 cases. Note also that COMBO outperforms all other methods on FlyingChairs.

%Finally, in Table \ref{tab:superv3}, we show the comparison of methods trained on the full curriculum with training data from FlyingChairs, FlyingThings, Sintel, HD1K and KITTI, evaluated on the Sintel and KITTI test sets (online benchmark). We remark that COMBO improves over RAFT on Sintel-clean, but is slightly inferior on Sintel-final and KITTI. COMBO is still superior to other methods on FlyingChairs, showing that the performances of COMBO deteriorate less when finetuning to a specific target dataset.
%In the next section, we show that COMBO reaches similar performances from earlier checkpoints in the semi-supervised setting.

%\begin{figure*}
%    \centering
%    \includegraphics[width=\linewidth]{latex/images/image_600.png}
%    \caption{Caption}
%    \label{fig:my_label}
%\end{figure*}

\subsection{Semi-supervised results}

\begin{table}[t]
    \centering
    \begin{tabular}{ccccc}
    \toprule
    model & mode  & checkpoint      &  \multicolumn{2}{c}{KITTI (test-resplit)} \\
    \cmidrule{4-5}
    ~ & ~ & ~ & F1-epe & F1-all \\
         \midrule
 %         RAFT & sup & None &  2.62 & 9.08 \\
%         COMBO & sup & None  & 6.71 & 27.3  \\        
         RAFT \cite{raft} & sup & C & 2.28 & 7.60 \\
         COMBO & sup & C & 2.19  & 7.30 \\
 COMBO & semisup  & C    & \textbf{1.75} & \textbf{6.88}  \\     
        \midrule
         RAFT \cite{raft} &  sup & S &  1.79 & 7.12  \\
        COMBO & sup & S &  1.75 & \textbf{6.58} \\ 
%  COMBO &   semisup  & T &  1.68 & 7.21  \\
 COMBO &    semisup &  S &  \textbf{1.74} & \textbf{6.58}  \\
     \bottomrule
    \end{tabular}
    \caption{Comparison with RAFT on KITTI-test-resplit from the FlyingChairs (C) and Sintel (S) checkpoints, with supervised and semisup training on KITTI. We see that COMBO in semisup from Chairs is almost equivalent to COMBO from Sintel, showing that the training curriculum can be drastically reduced.}
    \label{tab:semisup}
\end{table}

We analyze the performances of COMBO in a semi-supervised learning setting in Table \ref{tab:semisup}. We evaluate the different models on KITTI test-resplit, using for training the annotated images of KITTI train-resplit (sup) and the additional unlabelled images of KITTI (semisup). When training on KITTI from the earliest stage of the curriculum (FlyingChairs), we can see that COMBO-semisup (Fl-epe=1.75) largely improves over RAFT-sup (2.28) and COMBO-sup (2.19). It confirms that leveraging the vast amount of unlabelled frames is crucial for adapting the model to a target dataset from an early checkpoint. Interestingly, the semisup training from Chairs is almost equivalent to the COMBO model trained from the Sintel checkpoint (1.74). It shows that COMBO trained in a semisup context enables to greatly simplify the training curriculum to reach similar performances. This feature of COMBO is of crucial interest in many domains for transferring models to target datasets from a single synthetic training stage.

%\begin{minipage}{0.65\linewidth}
%\hspace{-0.3cm}
%To further analyze the semi-supervised learning ability of COMBO, we progressively reduce the number of training images on KITTI-resplit and compare the performances of RAFT and COMBO-semisup trained from the FlyingChairs checkpoint. In Figure \ref{fig:percentage}, we observe that RAFT degrades much more sharply than COMBO with fewer training images, and the difference in EPE becomes very large with 10\% of training images (EPE=11.6 for RAFT v.s. 4.4 for COMBO). It highlights the crucial ability of COMBO to leverage the BC assumption on unlabelled frames. 
%\end{minipage} %%
%\begin{minipage}{0.05\linewidth}
%\end{minipage} %%
%\begin{minipage}[c]{0.4\linewidth}
%\centering
%  \includegraphics[width=5cm]{images/pourcentage.png}
% \centering
% \captionof{figure}{Performances on KITTI-resplit trained from  Chairs, w.r.t. the number of labelled frames.}
%  \label{fig:percentage}
%\end{minipage}

%\vspace{-0.5cm}
\begin{wrapfigure}{r}{0.46\textwidth}
\vspace{-1.3cm}
  \begin{center}
   \includegraphics[width=5.3cm]{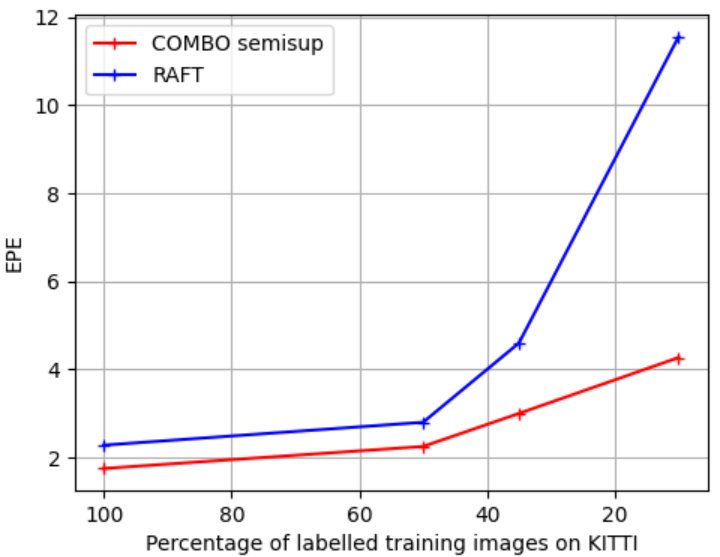}
  \vspace{-0.3cm}
  \caption{Performances on KITTI-resplit trained from the Chairs, w.r.t. the number of labelled frames.}
 \label{fig:percentage}
    \end{center}
\end{wrapfigure}
To further analyze the semi-supervised learning ability of COMBO, we progressively reduce the number of training images on KITTI-resplit and compare the performances of RAFT and COMBO-semisup trained from the FlyingChairs checkpoint. In Figure \ref{fig:percentage}, we observe that RAFT degrades much more sharply than COMBO with fewer training images, and the difference in EPE becomes very large with 10\% of training images (EPE=11.6 for RAFT v.s. 4.4 for COMBO). It highlights the crucial ability of COMBO to leverage the BC assumption on unlabelled frames.

\begin{figure*}
    \centering
    \includegraphics[width=\linewidth]{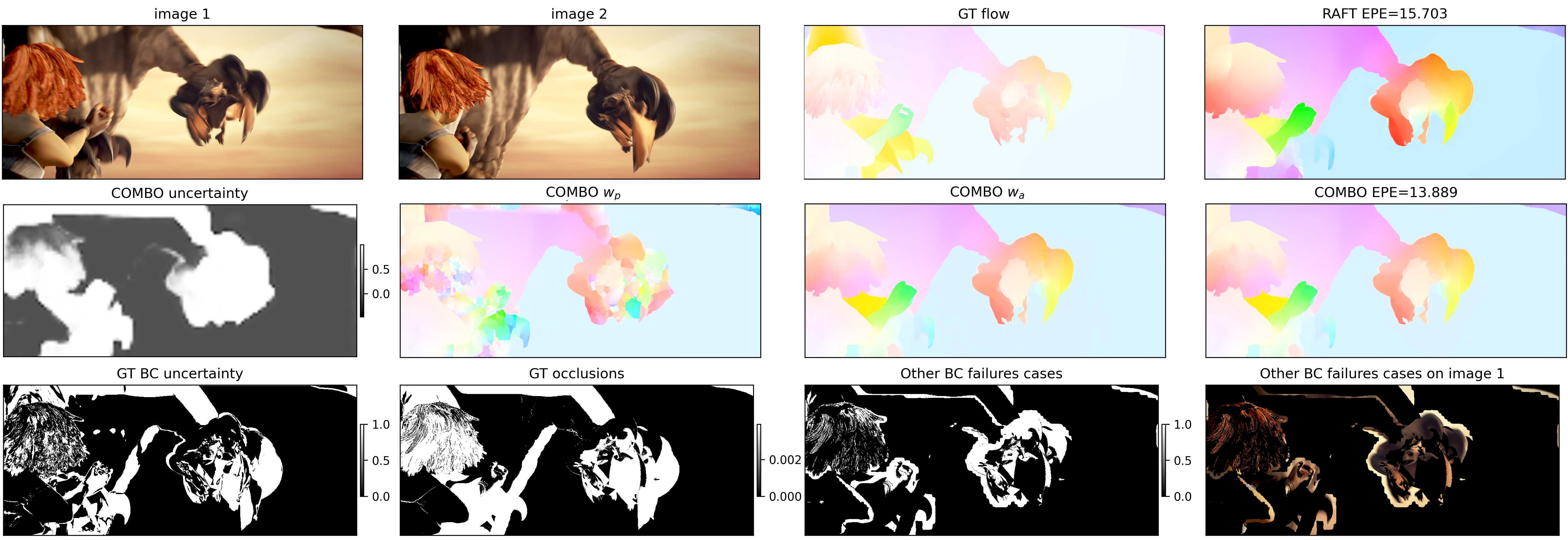}
    \caption{Qualitative prediction results on MPI Sintel. We observe that in zones of high uncertainty of the BC, the physical flow $\hat{\w}_p$ is ill-defined; the augmentation flow $\hat{\w}_a$ successfully complements it for accurate flow estimation. We also show that the uncertainty learned by COMBO is consistent with the GT BC uncertainty, and that it detects other cases of violation of the BC different from occlusions.}
    \label{fig:visu_sintel}
\end{figure*}

\begin{figure*}
    \centering
    \includegraphics[width=\linewidth]{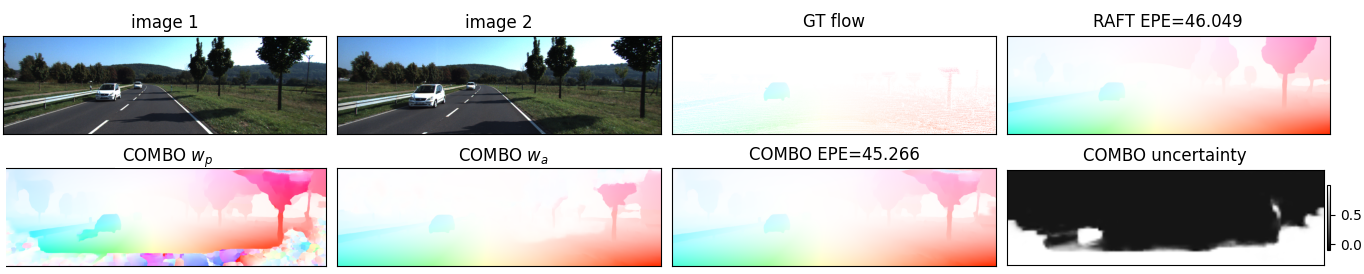}
    \caption{Qualitative prediction results on KITTI 2015. We see that at the bottom of the image where the road becomes occluded, the physical flow $\hat{\w}_p$ is ill-defined and is complements by $\hat{\w}_a$ for accurate prediction.}
    \label{fig:visu_kitti}
\end{figure*}

\begin{figure*}
    \centering
    \includegraphics[width=\linewidth]{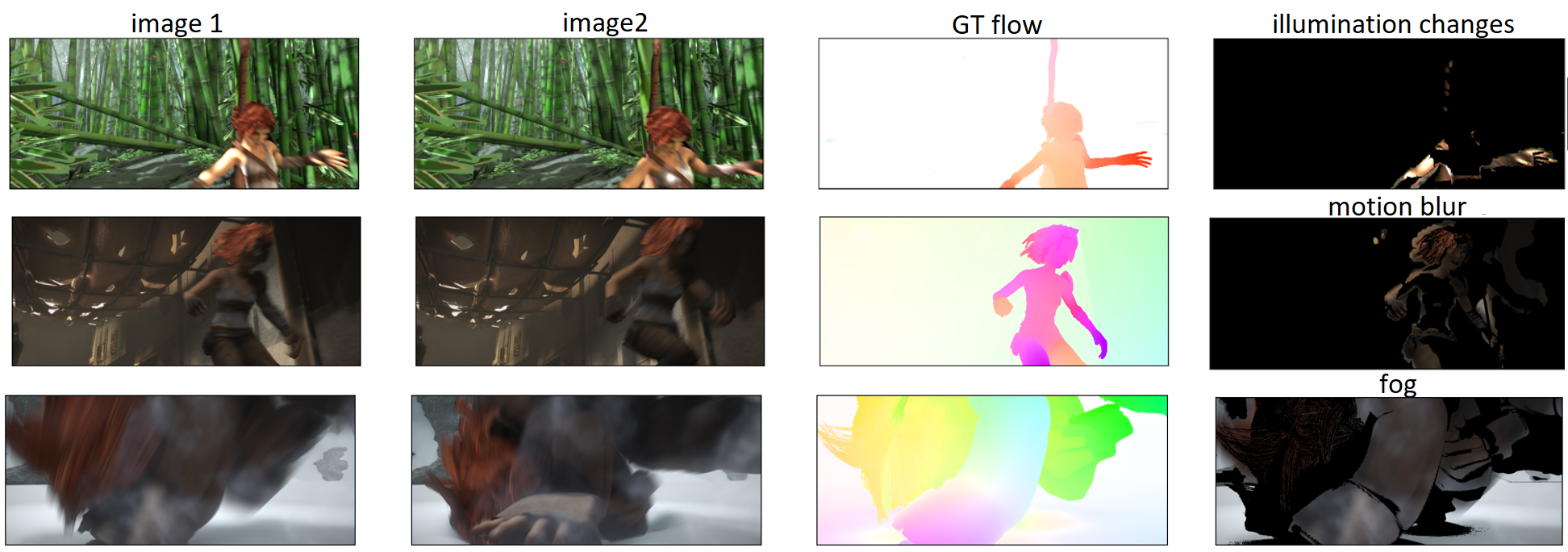}
    \caption{Failure cases of the brightness constancy detected by COMBO different from occlusions. The last column corresponds to image 1 masked by the map of thresholded uncertainty $\hat{\alpha}$ deprived of the ground-truth occlusions. From top to bottom, COMBO captures: illumination change, motion blur and fog.}
    \label{fig:failure_cases}
\end{figure*}

\subsection{Ablation study}

We perform in Table \ref{tab:ablation1} an ablation study on FlyingChairs to analyze the behaviour of the COMBO model. We first show that two RAFT branches trained with different seeds and mixed 50:50 improve over RAFT but is largely inferior to COMBO, show that the performances of COMBO do not simply come from an augmentation of the parameters or an ensembling effect. Then, we see that a single RAFT branch with the photometric loss $\mathcal{L}_{photo}$ in Eq. (\ref{eq:l_photo}) (epe=0.794) is far inferior to COMBO (0.743). This is because the BC regularization helps in improving generalization in regions where the BC assumption is fulfilled, but is detrimental in regions where the BC assumption does not hold. In contrast, COMBO can leverage the BC assumption and complement it in an adaptive manner when it is violated, overcoming the conflict issues between the two terms. 

Moreover, the linear decomposition $\hat{\w}=\hat{\w}_p+\hat{\w}_a$ without the uncertainty branch, \ie we set $\alpha(\x)=1/2$ everywhere, is clearly inferior to COMBO. It shows the crucial necessity to account for the uncertainty of the BC. Finally, when we remove the minimization of $\Vert (\hat{\w}_a, \hat{\w}_p) \Vert$ from COMBO (then the decomposition is not unique anymore), the performances also decrease, highlighting  the superiority of the well-posed and unique decomposition of COMBO.

We then report in Table \ref{tab:ablation2} the performances of the individual $\hat{\w}_p$ (resp. $\hat{\w}_a$) branches weighted by the certainty of the BC (resp. uncertainty). For example, we compute  $EPE_{weighted}(\hat{\w}_p, \w^*) = \sum_{\x} (1-\alpha^*(\x)) \Vert \hat{\w}_p(\x) - \w^*(\x) \Vert_2 $. We see that both $\hat{\w}_p$ and $\hat{\w}_a$ are superior to RAFT on areas where they apply. In particular the performance gap compared to RAFT widens on areas with high BC uncertainty (EPE= 2.20 \textit{v.s.} 0.52), showing the benefits of the COMBO decomposition to greatly improve the estimation on challenging zones.

%which are individually inferior but which successfully cooperate when adequately combined with the uncertainty $\alpha$, leading to a boost of performances.

\begin{table}[]
    \centering
    \begin{tabular}{cc}
    \toprule
    Method & EPE (FlyingChairs) \\
         \midrule
         RAFT \cite{raft} & 0.818 \\
         2 RAFT branches mixed equally & 0.800 \\ 
RAFT + BC photometric loss & 0.794  \\
     COMBO   $\hat{\w}_p+\hat{\w}_a$ (no weighting by uncertainty) & 0.769 \\
COMBO without $\min \Vert (\hat{\w}_a, \hat{\w}_p)  \Vert$  & 0.752 \\    
 COMBO &  \textbf{0.743} \\
     \bottomrule
    \end{tabular}
    \caption{Ablation study on FlyingChairs, showing the benefits of the meaningful and principled decomposition of COMBO, compared to the backbone RAFT.}
    \label{tab:ablation1}
\end{table}

\begin{table}[]
    \centering
    \begin{tabular}{cc}
    \toprule
    Method & EPE weighted (FlyingChairs) \\
    \midrule
RAFT on BC certain zones & 0.48 \\
    COMBO $\hat{\w}_p$ on certain zones  & \textbf{0.30} \\
     \midrule
    RAFT on BC uncertain zones & 2.29 \\
    COMBO $\hat{\w}_a$ on uncertain zones & \textbf{0.52}   \\
    
     \bottomrule
    \end{tabular}
    \caption{Performances of the individual BC $\hat{\w}_p$ and residual $\hat{\w}_a$ flow branches on areas with high BC confidence (resp. high BC uncertainty).}
    \label{tab:ablation2}
\end{table}

%Two sets of ablations: 
%\begin{itemize}  
%	\item High level ablations: RAFT, RAFT with a regularization wp single brach, COMBO wp, COMBA wa
%	\item Finer grained ablation: impact of our precise decomposition: RAFT, (wp, wa), (wp, wa, $\alpha$), COMBO
%\end{itemize}

\vspace{-0.5cm}
\subsection{COMBO analysis}

We provide qualitative flow predictions of the COMBO model for Sintel-test-resplit in Figure \ref{fig:visu_sintel} and KITTI-test-resplit in Figure \ref{fig:visu_kitti}. In uncertain zones ($\hat{\alpha} \approx 1$, for example in the occluded region of the bird in Figure  \ref{fig:visu_sintel} or at the bottom of Figure \ref{fig:visu_kitti}), we can see that $\hat{\w}_p$ gives incoherent results, as expected. In that case, the flow is efficiently complemented by  $\hat{\w}_a$ to produce the COMBO flow. 

\vspace{-0.2cm}
\textbf{Occlusion detection:} To analyze the ability of COMBO to detect occlusions, we represent in supplementary \ref{sup:combo-analysis} the precision-recall curve of the COMBO uncertainty detector with respect to the ground-truth occlusion masks (which are known for Sintel). COMBO obtains an Average Precision (AP) of 59\%, which is a lower bound of the true AP of COMBO for occlusions (since COMBO also captures other failure cases than occlusions). This means that COMBO is able to efficiently detect occlusions without any ground-truth occlusion supervision, compared to the random classifier which reaches 7\% (ratio of occluded pixels).

 \textbf{Detection of other failure cases of the BC:}
In the third line of Figure \ref{fig:visu_sintel}, we show the ground-truth (GT) BC uncertainty mask (defined as $\mathcal{L}_{BC} > \epsilon$, $\epsilon=0.01$ in Figure \ref{fig:visu_sintel}) and the ground-truth occlusions of Sintel. We observe that the COMBO uncertainty globally aligns with the GT BC uncertainty, and covers a larger area than the GT occlusions. To investigate this, we display the thresholded uncertainty mask $\alpha$ deprived of the GT occlusions: it shows that COMBO can capture other failure cases of the BC different than occlusions. Applying this mask on image 1 (bottom right of Figure \ref{fig:visu_sintel}) shows that this corresponds to illumination changes for this example.

We further analyze the other failure cases of the BC in Figure \ref{fig:failure_cases}. We can see that COMBO can detect other cases of violation of the BC, without any supervision masks on these cases. The top row of Figure \ref{fig:failure_cases} shows that the uncertainty $\alpha$ detects the illumination change on the arm of the woman. On the middle row, the uncertainty concentrates on the woman running, which causes motion blur. The bottom row shows a BC uncertainty on a large part of the image, caused by the presence of fog.
We provide additional visualizations in supplementary \ref{sup:visus}.

\vspace{-0.2cm}
\section{Conclusion}
\vspace{-0.2cm}
We have introduced COMBO, a new physically-constrained architecture for optical flow estimation that explicitly exploits the simplified brightness constancy constraint. COMBO learns the uncertainty of the BC and how to complement it with a data-driven network. COMBO reaches state-of-the-art results on several benchmarks, and is able to greatly simplify the training curriculum with a semi-supervised learning scheme. An appealing perspective would be to investigate the application of augmented physical models to multi-frame optical flow estimation.

%perspectives: extension to multi-frame

%plan d'expériences:

%\textbf{supervisé:}
%\begin{itemize}
%    \item ablation avec $w=w_p+w_a$ sans incertitude
%    \item ablation sans minimiser la norme de $w_a$ 
%    \item performances de la branche wp et wa seule
%    \item autre archi sur Chairs par ex PWNet
%    `\item analyse qualitative sur les cas d'échec: montrer la decompo, $\alpha$ etc pour occlusion, changement locaux globaux illu, reflexions speculaires etc
%\end{itemize}

%\textbf{semi-supervised:}
%\begin{itemize}
%    \item KITTI-raw + KITTI-resplit from scratch, à comparer avec KITTI-resplit from scratch\\
%    \item KITTI-raw + KITTI-resplit depuis les checkpoints chairs, things, Sintel-resplit et KITTI-resplit, à comparer avec les modèles supervisés équivalents sans les images non annotées
%    \item Chairs-sup + KITTI-raw (sans annotations sur KITTI), à comparer avec des méthodes unsupervised sur KITTI. \textcolor{red}{On risque de nous dire que c unfair / unsupervised ? Compa avec un modèle issu de Chairs-sup en transfert pur ? (ligne 1 tab 1 ) ou autre dans la même veine}
%    \item ablation sur le block gradient sur 1 ou plusieurs des cas ci-dessus
%    \
%\end{itemize}

\clearpage
\bibliographystyle{splncs04}
\bibliography{refs}
\end{document}

% \linenumbers
\pagestyle{headings}
\mainmatter
\def\ECCVSubNumber{100}  % Insert your submission number here

\title{Complementing Brightness Constancy with Deep Networks for Optical Flow Prediction \\ Supplementary material}

%\begin{comment}
%\titlerunning{ECCV-22 submission ID \ECCVSubNumber} 
%\authorrunning{ECCV-22 submission ID \ECCVSubNumber} 
%\author{Anonymous ECCV submission}
%\institute{Paper ID \ECCVSubNumber}
%\end{comment}
%******************

%\begin{comment}
\titlerunning{COMBO: Complementing Brightness Constancy for Optical Flow}
\author{Vincent Le Guen\inst{1,2,3} \and
Clément Rambour\inst{2} \and Nicolas Thome\inst{2,4}}

\authorrunning{Vincent Le Guen, Clément Rambour and Nicolas Thome}

\institute{EDF R\&D, Chatou, France \\
 \and
Conservatoire National des Arts et Métiers, CEDRIC, Paris, France \\ \and
SINCLAIR AI Lab, Palaiseau, France
 \and
Sorbonne Université, CNRS, ISIR, F-75005 Paris, France
}

%******************
\maketitle

\section{Proof of the uniqueness of the COMBO decomposition}
\label{sup:proof}

%\begin{align}
%\underset{\w_p, \w_a}{\min} ~~~  \left\Vert  (\w_a,\w_p)  \right\Vert ~~~~~~~ %\text{subject to: ~~~~~~}     \label{eq:optpb} \\
%\begin{cases}
%(1-\alpha(\x)) ~ \w_p(\x) + \alpha(\x) ~ \w_a(\x) = \w^*(\x)  \\
% (1-\alpha(\x))~ | I_1(\x) - I_2(\x+\w_p(\x))  | = 0  \nonumber   \\
% \alpha(\x) = \sigma \left( |I_1(\x)-I_2(\x+\w^*(\x))| \right).   \\
%\end{cases}
%\end{align}

We recall here the COMBO decomposition of the ground truth flow vector $\w^*(\mathbf{x})$ between a physical part $\w_p^*(\mathbf{x})$ fulfilling the Brightness Consistency (BC) assumption, an augmented term $\w_a^*(\mathbf{x})$, and a BC uncertainty term $\alpha^*(\mathbf{x})$:
\begin{equation}
   \w^*(\x)= (1-\alpha^*(\x)) ~ \w_p^*(\x) + \alpha^*(\x) ~ \w_a^*(\x).
   \label{eq:decomposup}
\end{equation}

Since the decomposition in Eq.~(\ref{eq:decomposup}) is not necessarily unique, the COMBO decomposition ($\w_p^*, \w_a^*, \alpha^*$) is defined as the solution of the following constrained optimization problem:

\begin{align}
\underset{\w_p, \w_a}{\min} ~~~  \left\Vert  (\w_a,\w_p)  \right\Vert ~~~~~~~ \text{subject to: ~~~~~~}     \label{eq:optpbsup} \\
\begin{cases}
(1-\alpha^*(\x)) ~ \w_p(\x) + \alpha(\x) ~ \w_a(\x) = \w^*(\x)  \\
 (1-\alpha^*(\x))~ | I_1(\x) - I_2(\x+\w_p(\x))  | = 0  \nonumber   \\
 \alpha^*(\x) = \sigma \left( |I_1(\x)-I_2(\x+\w^*(\x))| \right).   \\
\end{cases}
\end{align}

%\begin{align}
%  \w(\mathbf{x}) &= (1-\alpha(\mathbf{x})) ~ \w_p(\mathbf{x}) + \alpha(\mathbf{x}) ~ \w_a(\mathbf{x}) \text{~~s.t.:~~~} \nonumber \\
%    & \begin{cases}
%    \alpha(\mathbf{x}) = \sigma( \mathcal{L}_{BC}(\x,\w) ) \\
%  \w_p = arg~min_{\w_{p'}}  \mathcal{L}_{BC}(\x,\w_p') \\ 
%  \w_a = arg~min_{\w_{a'}} ||\w_a'||^2 \\
%    \w_p = arg~min_{\w_{p'}} ||\w_p'||^2 ,
%   \end{cases}
%   \label{eq:decomp}
%\end{align}

\begin{figure}[h]
     \centering
     \includegraphics[width=10cm]{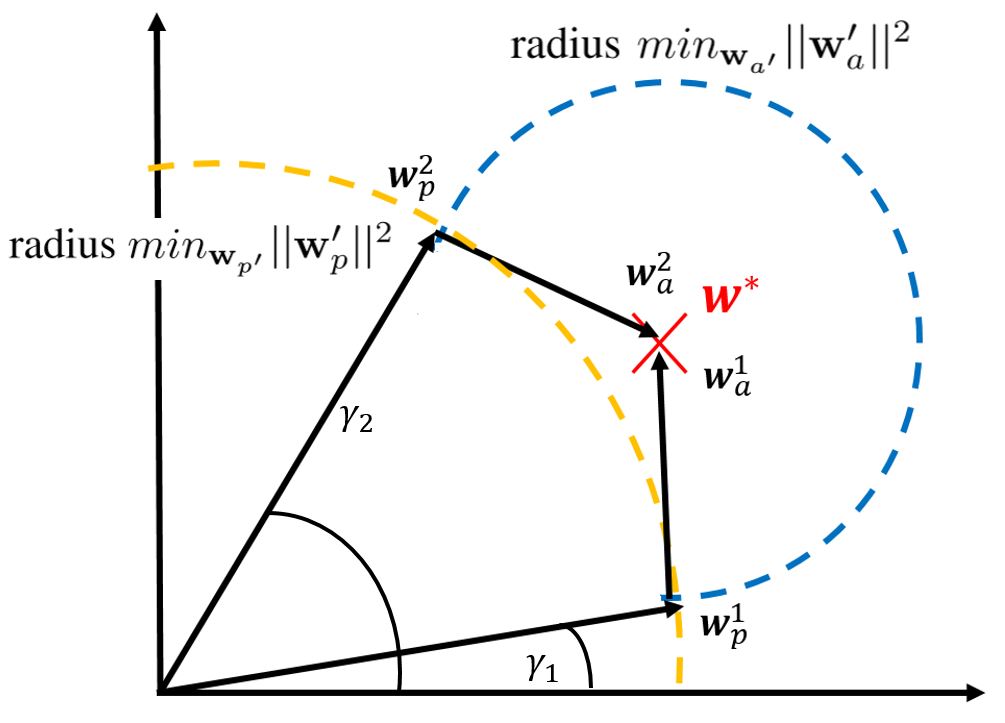}
     \caption{Illustration of the unique decomposition of COMBO.}     
    % \caption{Illustration of the unique decomposition of COMBO. $\alpha(\mathbf{x})$ is designed in an unique fashion: $\alpha(\mathbf{x}) = \sigma( \mathcal{L}_{BC}(\x,\w) )$. Then, the constraint $\w_p = arg~min_{\w_{p'}}  \mathcal{L}_{BC}(\x,\w_p')$ in \cref{eq:decomp} specifies a set of BC minimizers for $\w_p$. The orange (resp. blue) circle represents the $||\w_p||$ (resp. $||\w_a||$) constraint. The intersection between the two circle limits the solutions to the two vectors $(\w_p^1(\mathbf{x}),\w_a^1(\mathbf{x}))$ and $(\w_p^2(\mathbf{x}),\w_a^2(\mathbf{x}))$. By imposing the smallest $\w_p$ angle, we obtain the unique solution $(\w_p^1,\w_a^1,\alpha)$.}
     \label{fig:decomp}
 \end{figure}

We detail here the uniqueness guarantee of the COMBO decomposition in Eq.~(\ref{eq:optpbsup}). An unconstrained decomposition would be written as follows: 
\begin{equation}
  \w(\mathbf{x}) = (1-\alpha(\mathbf{x})) ~ \w_p(\mathbf{x}) + \alpha(\mathbf{x}) ~ \w_a(\x).
   \label{eq:decomp_naive}
\end{equation}  
It is clear that the naive decomposition in Eq.~(\ref{eq:decomp_naive}) admits multiple $(\w_p(\mathbf{x}),\w_a(\mathbf{x}),\alpha(\mathbf{x}))$ tuples. We highlight here the effect of the different constraints in Eq.~(\ref{eq:optpbsup})
:

\begin{itemize}
    \item $\alpha^*(\mathbf{x}) = \sigma( \mathcal{L}_{BC}(\x,\w) )$ specifies a unique value for $\alpha^*(\mathbf{x})$, but their remains an infinite number of $(\w_p(\mathbf{x}),\w_a(\mathbf{x}))$ tuples.
    \item $ (1-\alpha(\x))~ | I_1(\x) - I_2(\x+\w_p(\x))  | = 0$ specifies a set of BC minimizers for $\w_p$, that we denote $\mathcal{F}_{BC}$.
    \item By minimizing $||\w_p||$, we limit $\w_p(\mathbf{x}) \in \mathcal{F}_{BC}$ to obtain $\w_p(\mathbf{x}) \in \mathcal{F}_{BC} \cap C_p$, where $C_p$ is the circle of radius $min_{\w_{p'}} ||\w_p'||^2$ (orange in Fig. \ref{fig:decomp}). 
    
    \item By minimizing $||\w_a||$, we limit $(\w_p(\mathbf{x}),\w_a(\mathbf{x}))$ to the two sets  $(\w_p^1(\mathbf{x}),\w_a^1(\mathbf{x}))$ and $(\w_p^2(\mathbf{x}),\w_a^2(\mathbf{x}))$ shown in Fig. \ref{fig:decomp}, which is the intersection between the orange circle and the blue circle of radius $min_{\w_{a'}} ||\w_a'||^2$ in Fig. \ref{fig:decomp}. This constraint following the least action principle, and adds only the minimal information to the BC properly represent $\w(\mathbf{x})$. Finally, by minimizing the angle $\gamma = < \w_p; x >$, we obtain the unique solution $(\w_p^1(\mathbf{x}),\w_a^1(\mathbf{x}))$ shown in Fig \ref{fig:decomp}.
 \end{itemize}

%Minimizing the brightness constancy constraint $\mathcal{L}_{BC}$ leads to multiple possible solutions $\w_p$ since for a source pixel, there may exist multiple pixels in the target image with the same intensity.

%Following the least action principle, we further minimize the norm of the augmentation flow $\Vert \w_a \Vert$ so that it only captures the information that cannot be modelled by the physical prior. This gives another constraint that limits the number of possible solutions.

% However, the decomposition may still not be unique if there exists several physical flows $\w_p$ equidistant to $\w^*$ (see Figure \ref{fig:decompo} for an illustration). To overcome such a degenerate case, we select the $\w_p$ of minimal norm among the candidates, which gives the last constraint of our problem. 
 
%Finally the last degenerate case is if there exists only two $\w_p$ on the circle with the same norm (on each side of the vector $\w^*$).  In this case we can select heuristically the $\w_p$ with the minimal angle.
 
 Therefore \textbf{the decomposition in Eq. \ref{eq:optpbsup} admits a unique tuple $(\w_p^*,\w_a^*,\alpha^*)$}.%, with  $\w_p=\w_p^1(\mathbf{x})$, $\w_a=\w_a^1(\mathbf{x})$, and $\alpha=\sigma( \mathcal{L}_{BC}(\x,\w))$}.
 
\newpage
\section{Implementation details}
\label{sup:implem}

The code of COMBO is implemented in Pytorch. In the supervised setting, we train models with the following curriculum: FlyingChairs (100000 steps, learning rate=$4e-4$), FlyingThings3D (100000 steps, lr=0.000125), Sintel with additional data from HD1K, FlyingThings and KITTI (100000 steps, lr=0.000125), and finally KITTI (100000 steps, lr=0.001).
We use the Adam optimizer with a cyclic learning rate scheduler.

The hyperparameters choosen in the training loss are the following: $\lambda_{total}=1$, $\lambda_{p}=0.1$, $\lambda_{a}=0.01$, $\lambda_{photo}=0.01$, $\lambda_{\alpha}=1$, $\lambda_{w}=0.1$. This ensures a proper scaling between losses, but we do not attempt at finely tuning the hyperparameters due to the huge computational training time. This setting is kept constant for all stages of the curriculum and datasets.

Each step of the curriculum takes approximately 5 days to converge on a DGX A100 gpu. This gives a very long training curriculum as a whole, highlighting the benefits brought up by the simplification with the semi-supervised training of COMBO.

%\begin{table}[h]
%   \centering
%   \begin{adjustbox}{max width=\linewidth}
%    \begin{tabular}{c|cccc}
%  \toprule
%Method &  Chairs & \multicolumn{2}{c}{Sintel (test)} & KITTI-15 (test)  \\
%\cmidrule(r){3-4}  \cmidrule(r){5-5}
% ~ & ~ & Clean & Final & Fl-all \\
% \midrule
%FlowNet 2.0 \cite{ilg2017flownet}  & - & 3.96 & 6.02 & 10.41 \\
%pyNet \cite{ranjan2017optical} & - &  6.64 & 8.36 & 35.07 \\
%LiteFlowNet2 \cite{hui2020lightweight} & -   & 3.48 & 4.69 & 7.74  \\
%     PWC-Net+ \cite{sun2019models} & 2.00  & 3.45 & 4.60 & 7.72  \\ 
%  VCN \cite{yang2019volumetric} & -  & 2.81 & 4.40 & 6.30  \\
% MaskFlowNet \cite{zhao2020maskflownet} & -  & 2.52 & 4.17 & 6.10  \\     
%  RAFT \cite{raft}  & 1.23  & 1.61 & 2.86 & \textbf{5.10}  \\
% C+T+S+H+K &    RAFT resplit & 2.24 & (1.95) & (4.77) & (1.17) & (4.05)  \\
% GMA \cite{jiang2021learning} & 1.27  & \textbf{1.39} & \textbf{2.47} & 5.15  \\
% COMBO  & \textbf{1.16}  & 1.60 & 3.41 & 6.03   \\
%   \bottomrule
%    \end{tabular}
%    \end{adjustbox}
%    \caption{State-of-the-art performance comparison on FlyingChairs and on the test set of Sintel and KITTI from the online benchmark. Models are trained on the full curriculum.}
%    \label{tab:superv3}
%\end{table}

\section{Experiments}

\subsection{Influence of the backbone model}

In the main paper, we validate the performances of COMBO based on the RAFT \cite{raft} backbone architecture, which is currently one of the state-of-the-art models. However, the COMBO rationale of leveraging the brightness constancy in a deep augmented model is agnostic to the backbone model. We conduct an additional experiment on top of the very recent GMA model \cite{jiang2021learning}. The results shown below (Table \ref{tab:gma}) on the test set prove that COMBO still provides a significant improvement (epe=0.71 \textit{v.s.} 0.82) compared to this state-of-the-art GMA model on the FlyingChairs stage, showing that COMBO is a general augmentation strategy for the BC, agnostic to the optical flow architecture.
\begin{table}[h]
    \centering
     \begin{adjustbox}{max width=\linewidth}
    \begin{tabular}{c|c}
    \toprule
        RAFT & 0.82 \\
        COMBO (backbone RAFT) &  \textbf{0.74} \\
        \midrule
        GMA & 0.82 \\
        COMBO (backbone GMA) &   \textbf{0.71} \\

    \bottomrule
    \end{tabular}
    \end{adjustbox}
     \vspace{0.2cm}
\caption{Performances of COMBO on the FlyingChairs dataset, based on the RAFT \cite{raft} and GMA \cite{jiang2021learning} backbone architectures.}
    \label{tab:gma}
\end{table}

\subsection{Examples of supervision $(\w_p^*$, $\w_a^*$, $\alpha^*$)}
\label{sup:supervision}

\begin{figure}
    \centering
    \includegraphics[width=\linewidth]{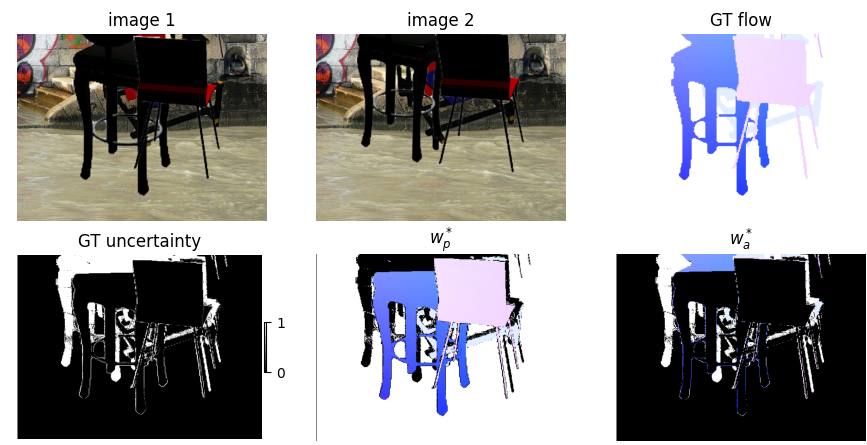}
    \caption{Example of ground truth supervision $(\w_p^*$, $\w_a^*$, $\alpha^*$) for FlyingChairs.}
    \label{fig:supervision_chairs}
\end{figure}

\begin{figure}
    \centering
    \includegraphics[width=\linewidth]{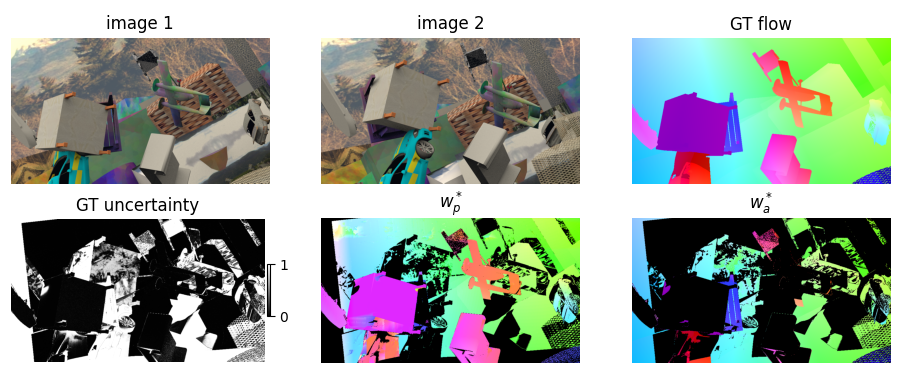}
  \caption{Example of ground truth supervision $(\w_p^*$, $\w_a^*$, $\alpha^*$) for FlyingThings3D.}
    \label{fig:supervision_things}
\end{figure}

\begin{figure}
    \centering
    \includegraphics[width=\linewidth]{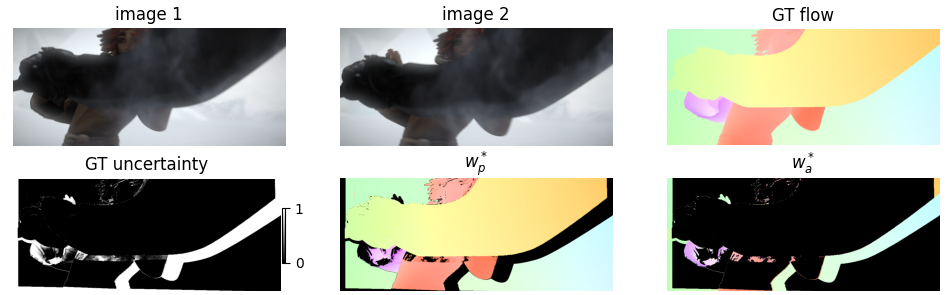}
  \caption{Example of ground truth supervision $(\w_p^*$, $\w_a^*$, $\alpha^*$) for Sintel.}
    \label{fig:supervision_sintel}
\end{figure}

\begin{figure}
    \centering
    \includegraphics[width=\linewidth]{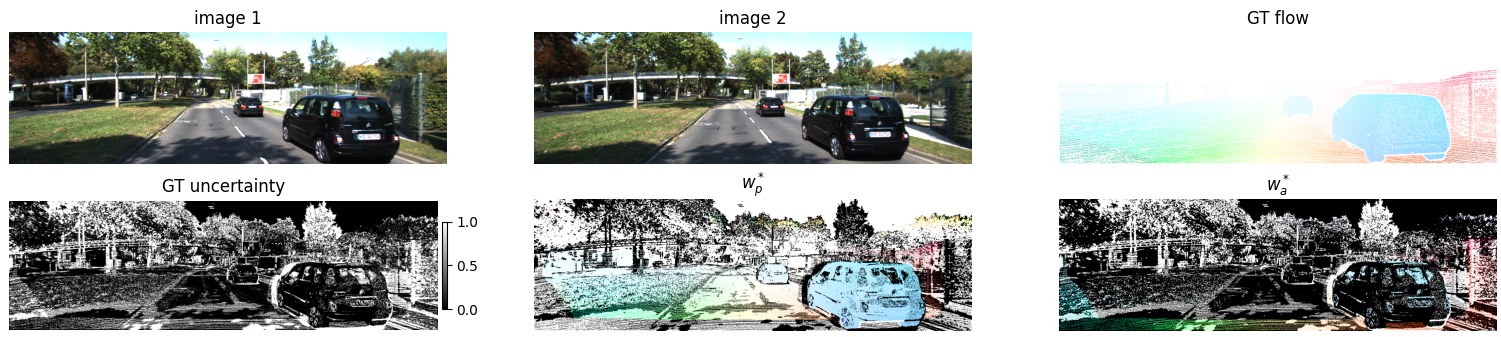}
  \caption{Example of ground truth supervision $(\w_p^*$, $\w_a^*$, $\alpha^*$) for KITTI.}
    \label{fig:supervision_kitti}
\end{figure}

We provide a few examples of ground truth supervision $(\w_p^*$, $\w_a^*$, $\alpha^*$) for the datasets FlyingChairs (Figure \ref{fig:supervision_chairs}), FlyingThings3D (Figure \ref{fig:supervision_things}), Sintel (Figure \ref{fig:supervision_sintel}) and KITTI (Figure \ref{fig:supervision_kitti}).

\subsection{COMBO analysis}
\label{sup:combo-analysis}

\begin{figure}[h]
    \centering
    \includegraphics[width=8cm]{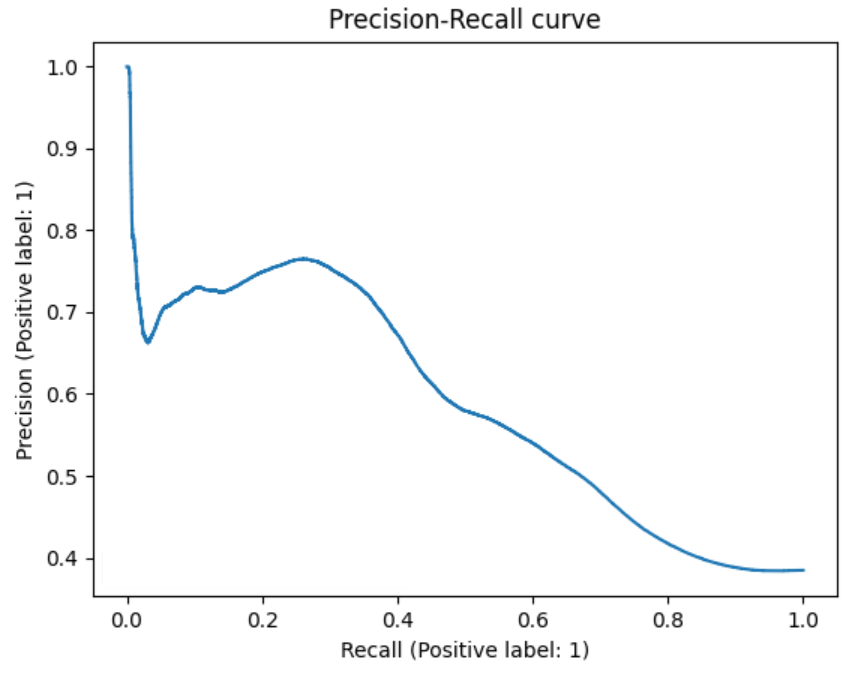}
    \caption{Precision-recall curve quantifying the ability of COMBO to detect occlusions.}
    \label{fig:PR_curve}
\end{figure}

To analyze the ability of COMBO to detect occlusions, we show in Figure \ref{fig:PR_curve} the precision-recall curve of the COMBO uncertainty detector with respect to the ground-truth occlusion masks (which ground truth is provided on Sintel). COMBO obtains an Average Precision (AP) of 59\% for occlusion detection. This average precision underestimates the true AP since COMBO detects occlusions and also other failure cases: therefore, the precision computed here penalizes a correct BC violation detection (large $\alpha$) which is not labeled as occlusion. Similarly, some occluded pixels fulfilled the BC assumption, making the recall computed in this manner overestimated.  Therefore, the AP reported here is a lower bound of the true AP for occlusions. Despite this, it shows that COMBO is able to efficiently detect occlusions without any ground-truth occlusion supervision, compared to the random classifier which reaches 7\% (ratio of occluded pixels).

\subsection{Additional visualizations}
\label{sup:visus}

\begin{figure*}
    \centering
    \includegraphics[width=\linewidth]{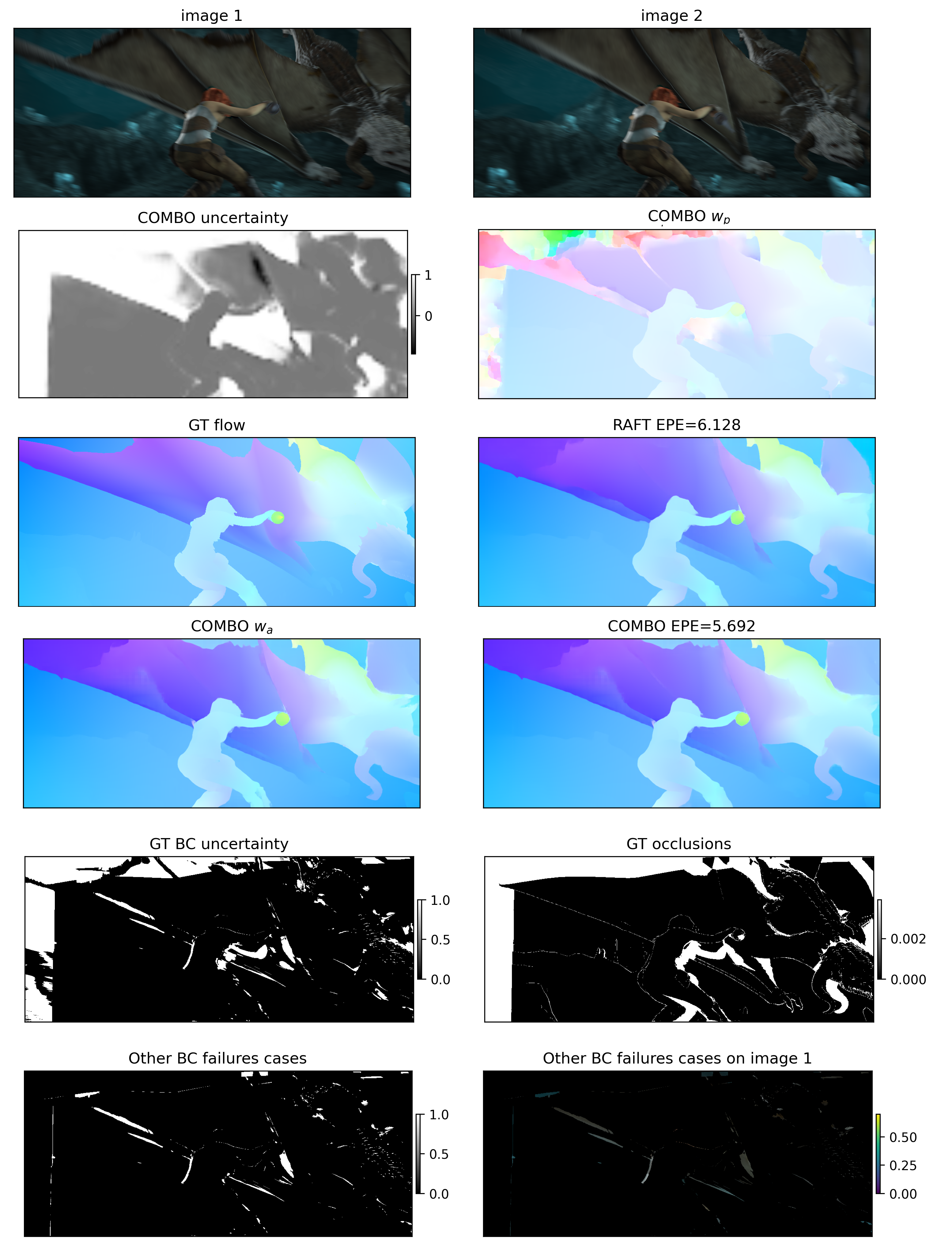}
      \caption{Additional visualization on Sintel.}
       \label{fig:sintel_supp1}
\end{figure*}

\begin{figure*}
    \centering
    \includegraphics[width=\linewidth]{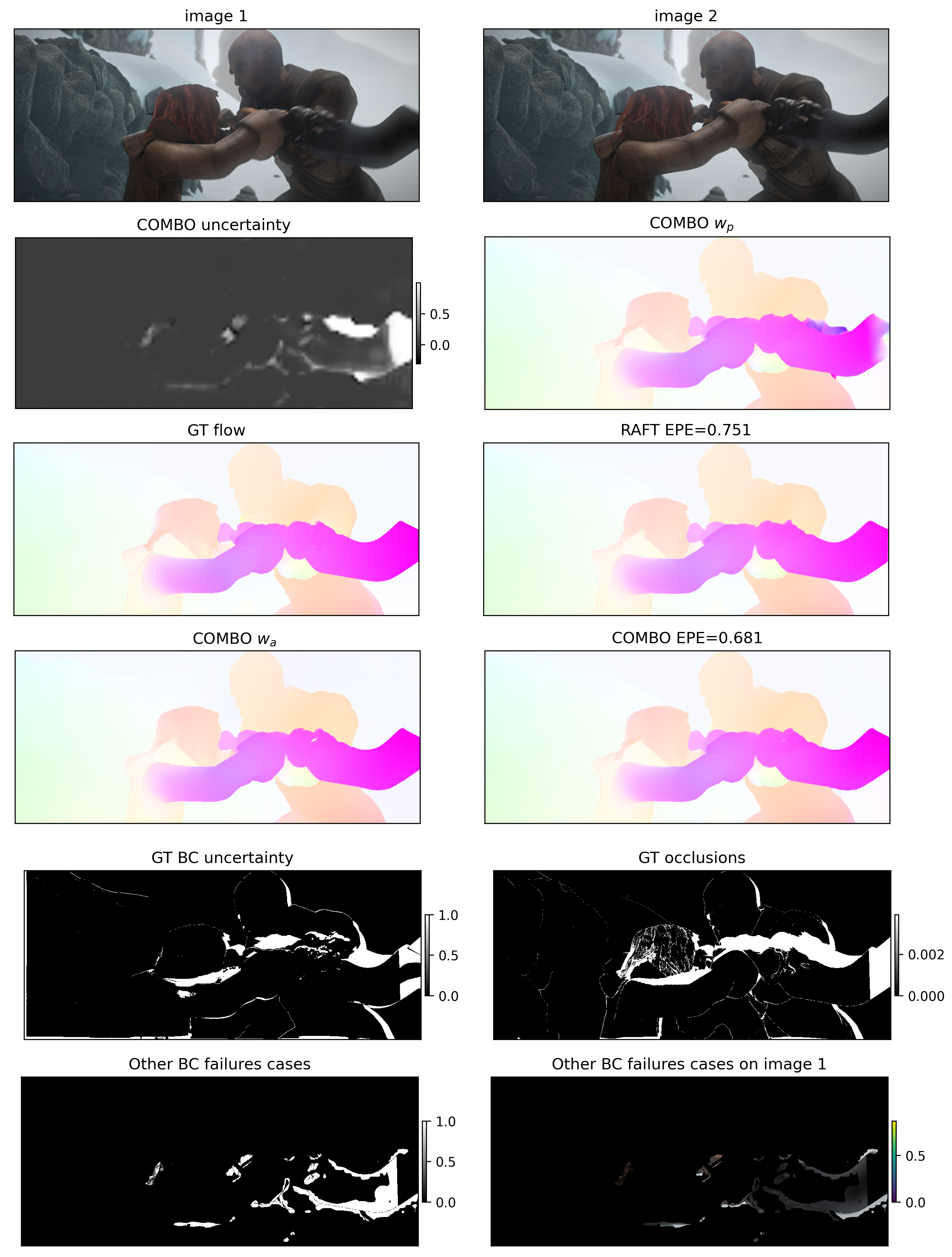}
      \caption{Additional visualization on Sintel.}
       \label{fig:sintel_supp2}
\end{figure*}

\begin{figure*}
    \centering
    \includegraphics[width=\linewidth]{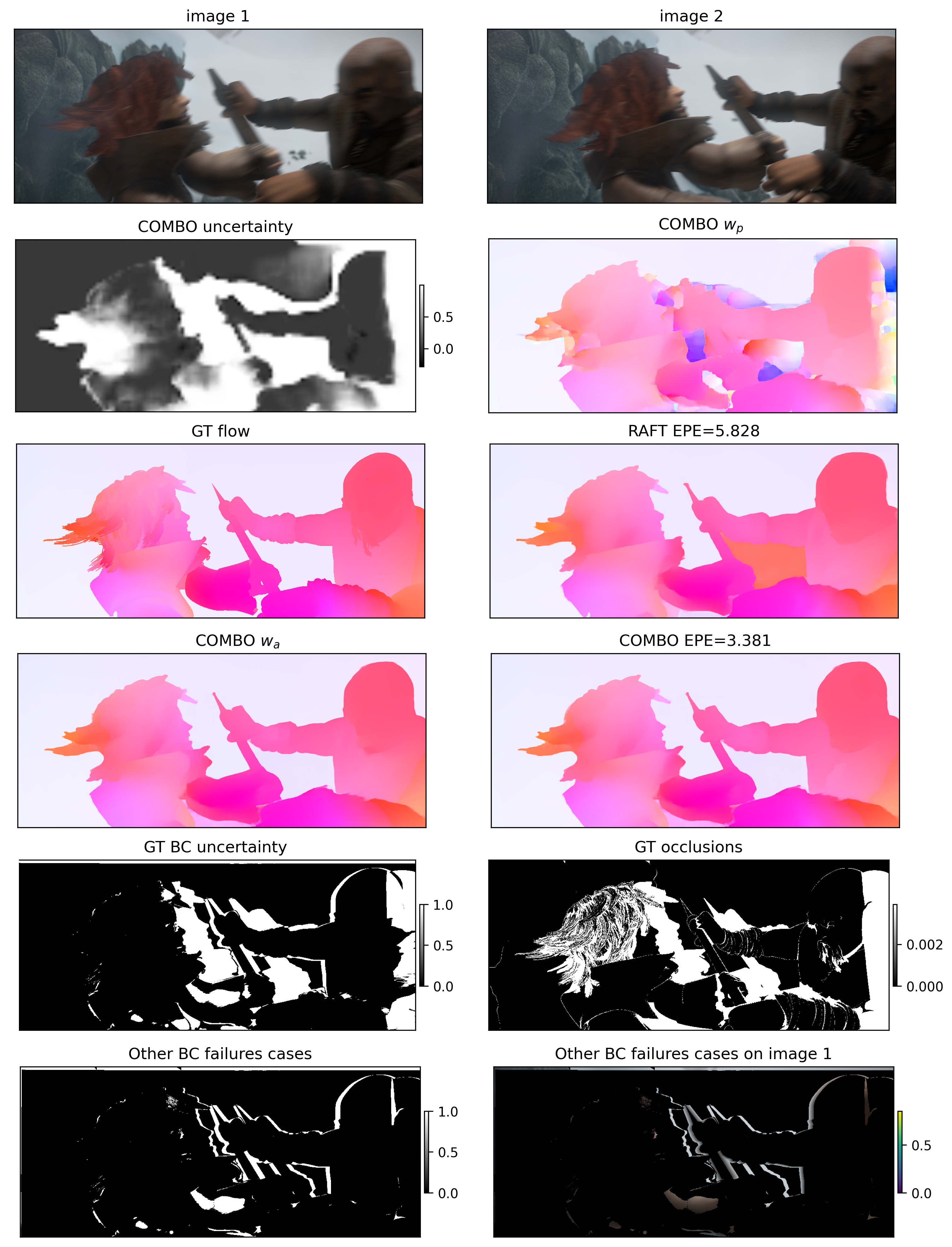}
    \caption{Additional visualization on Sintel.}
    \label{fig:sintel_supp3}
\end{figure*}

\begin{figure*}
    \centering
    \includegraphics[width=\linewidth]{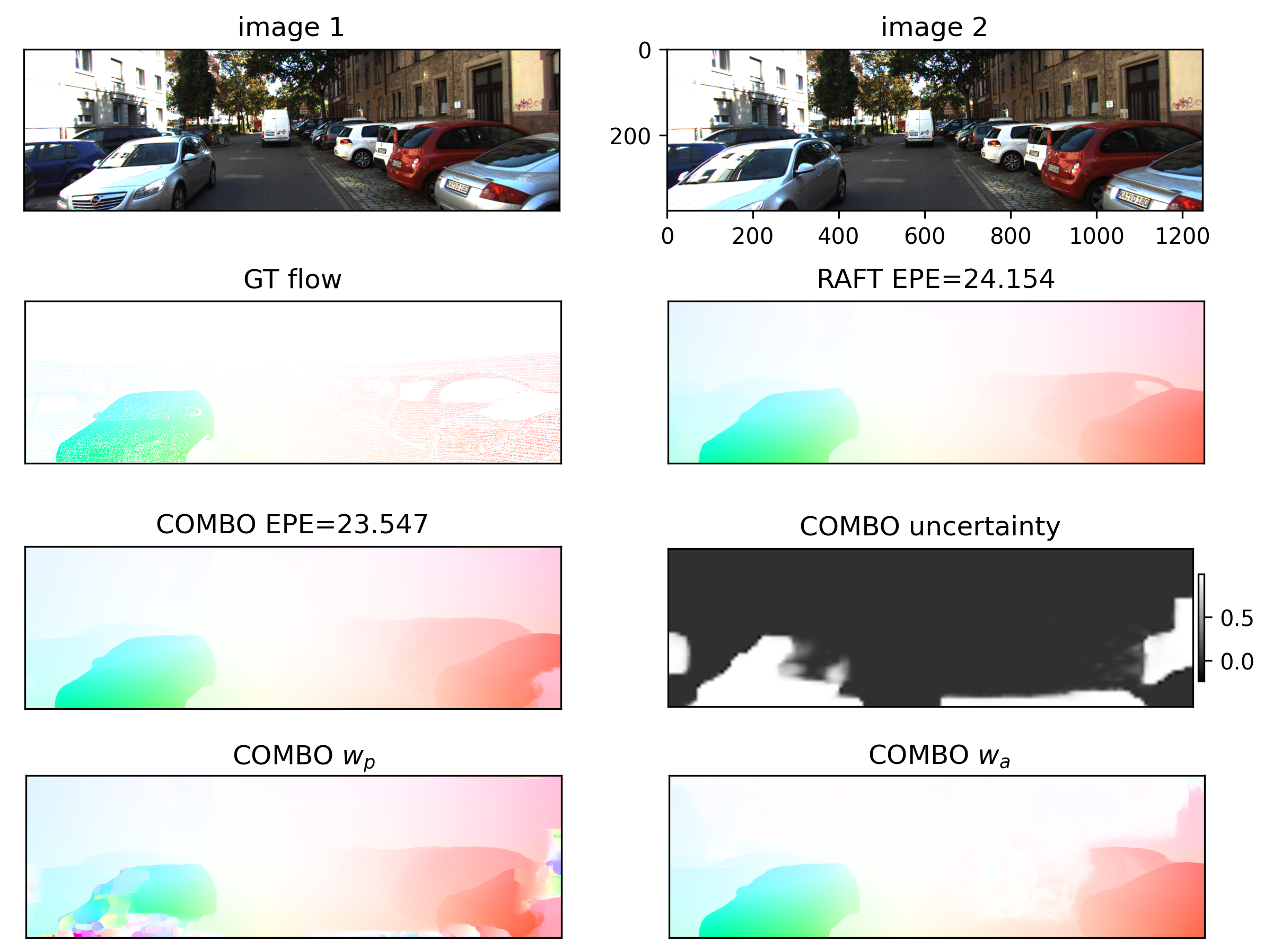}
        \caption{Additional visualizations on KITTI-2015.}
    \label{fig:kitti_supp1}
\end{figure*}

\begin{figure*}
    \centering
    \includegraphics[width=\linewidth]{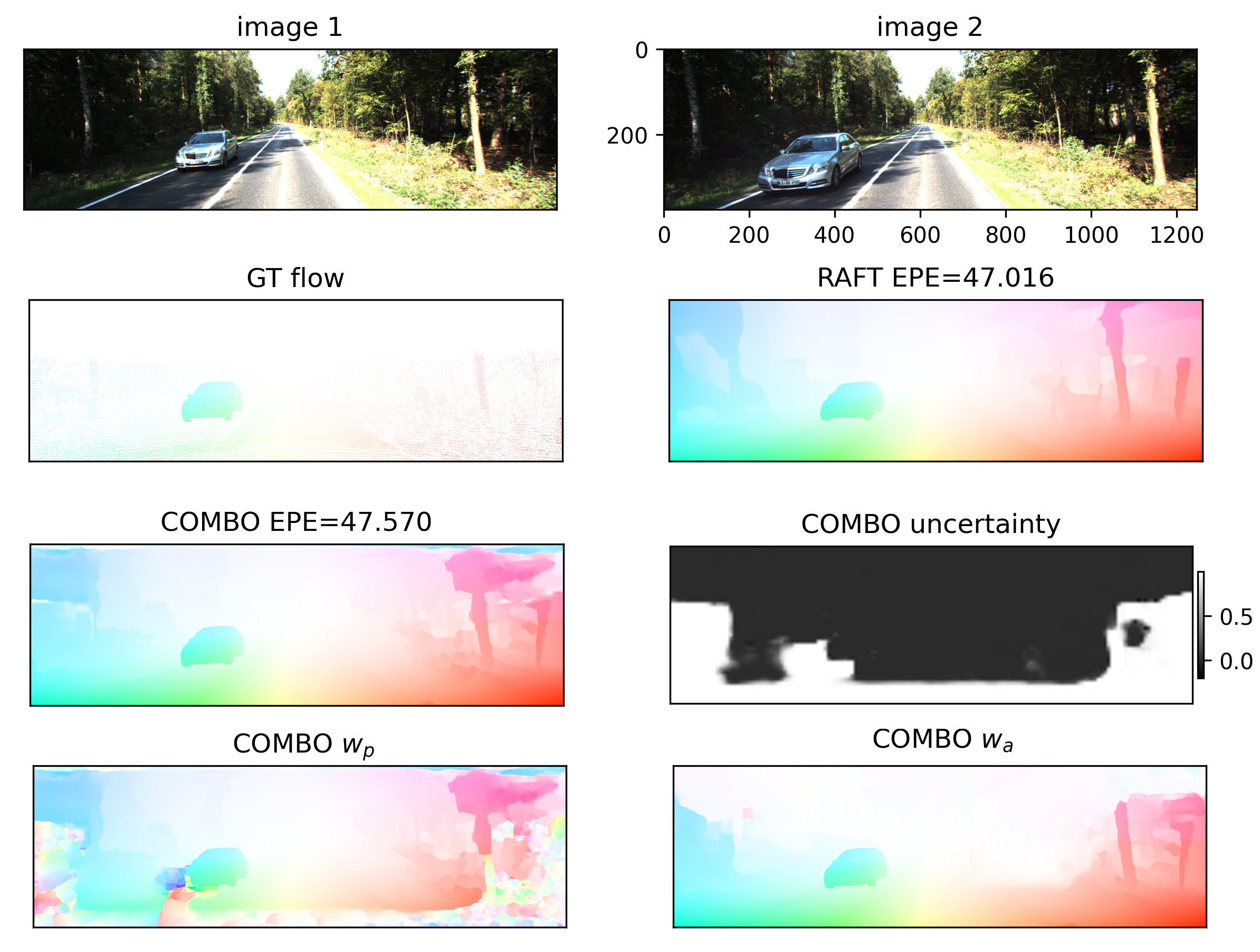}
        \caption{Additional visualizations on KITTI-2015.}
    \label{fig:kitti_supp2}
\end{figure*}

\begin{figure*}
    \centering
    \includegraphics[width=\linewidth]{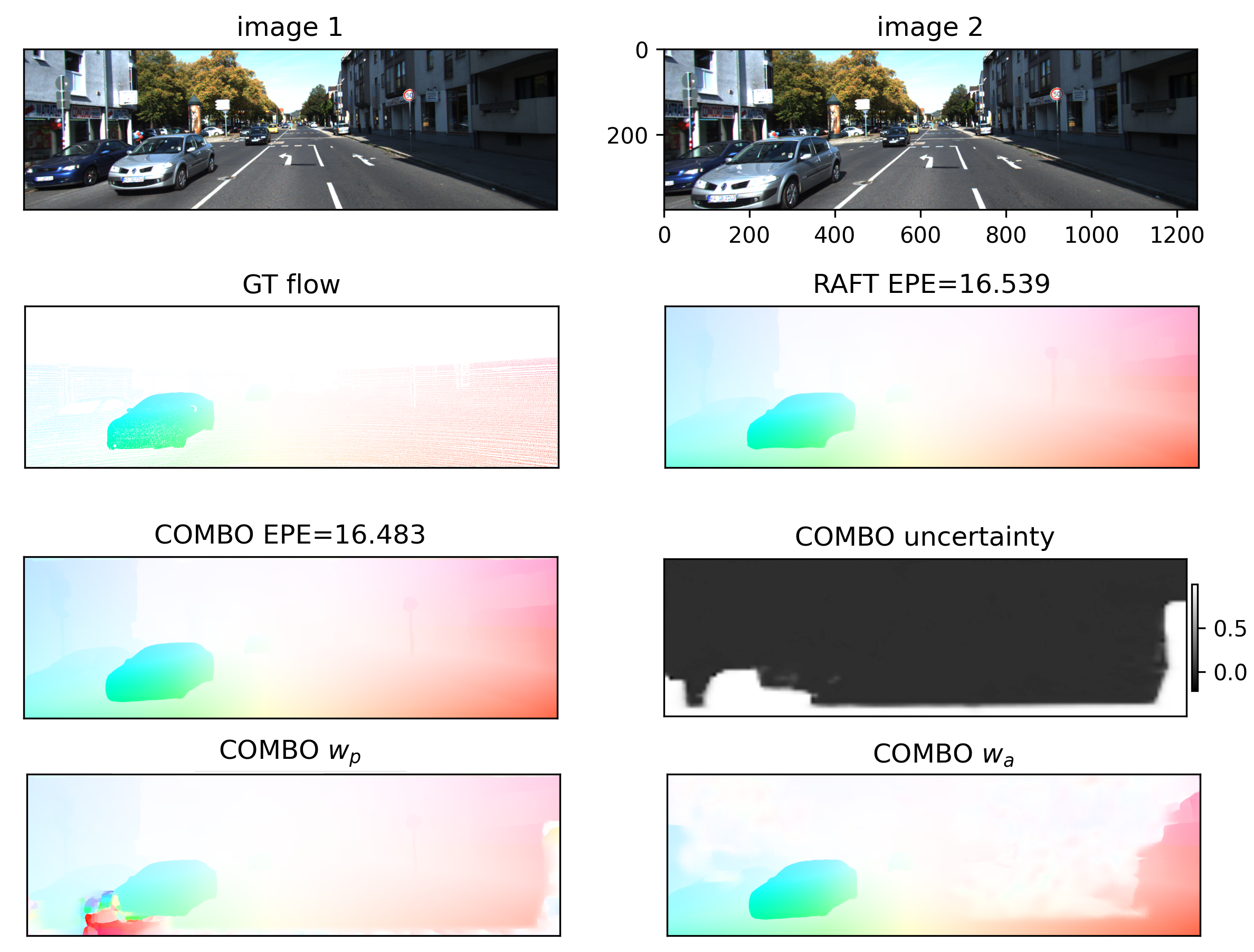}
    \caption{Additional visualizations on KITTI-2015.}
    \label{fig:kitti_supp3}
\end{figure*}

We provide additional visualizations for Sintel in Figures \ref{fig:sintel_supp1}, \ref{fig:sintel_supp2}, \ref{fig:sintel_supp3} and KITTI-2015 in Figures \ref{fig:kitti_supp1}, \ref{fig:kitti_supp2}, \ref{fig:kitti_supp3}.

In each case, we can observe that in zones of high uncertainty of the brightness constancy, the physical flow $\w_p$ is ill-defined; in these zones, it is efficiently complemented by the flow $\w_a$ to produce an accurate COMBO flow estimate.

\vspace{2cm}

\clearpage
\bibliographystyle{alpha}
\bibliography{egbib}